\RequirePackage{fix-cm}
\documentclass[10pt,letterpaper]{article}

\usepackage[textwidth=6.875in,textheight=9in,top=1in]{geometry}
\usepackage{times,xspace}
\usepackage[dvipsnames]{xcolor}
\usepackage{graphicx,amsmath,amssymb,booktabs}
\usepackage[format=plain,labelsep=period,font=small]{caption}
\usepackage[shortlabels,inline]{enumitem}
\usepackage[numbers,sort&compress]{natbib}
\usepackage[hyphens]{url}
\DeclareUnicodeCharacter{00B1}{\ensuremath{\pm}}
\DeclareUnicodeCharacter{2212}{\ensuremath{-}}
\DeclareUnicodeCharacter{2192}{\ensuremath{\to}}

\usepackage{array,tabularx}
\usepackage{multicol}
\usepackage{microtype}
\newcolumntype{Y}{>{\raggedright\arraybackslash}X}

\newcommand{\bench}{EVD-Bench\xspace}
\newcommand{\evdTableValue}[2]{\mbox{#1{\fontsize{6}{7}\selectfont\,$\pm$\,#2}}}

\renewcommand{\textfraction}{0.1}
\definecolor{linkblue}{rgb}{0.12,0.32,0.54}
\usepackage[colorlinks,allcolors=linkblue]{hyperref}
\newcommand{\PaperTitle}{EVD-Bench: A Comprehensive Benchmark for EEG-to-Image Decoding}
\newcommand{\PaperAuthors}{Kieren Yu, Qian Zhang, Ziyang Liu, Chang Huang, Jintai Chen, Kaishun Wu}
\hypersetup{pdftitle={\PaperTitle},pdfauthor={\PaperAuthors}}

\title{\PaperTitle}
\author{\PaperAuthors\\[4pt]
  \normalsize The Hong Kong University of Science and Technology (Guangzhou)}

\makeatletter
\renewcommand{\@maketitle}{%
  \begin{center}
    {\Large\@title\par}
    \vspace{10pt}
    {\large\@author\par}
  \end{center}
  \vspace{8pt}
  \begingroup
  \captionsetup{hypcap=false}%
  \begin{minipage}{\textwidth}
\centering
\includegraphics[width=\linewidth]{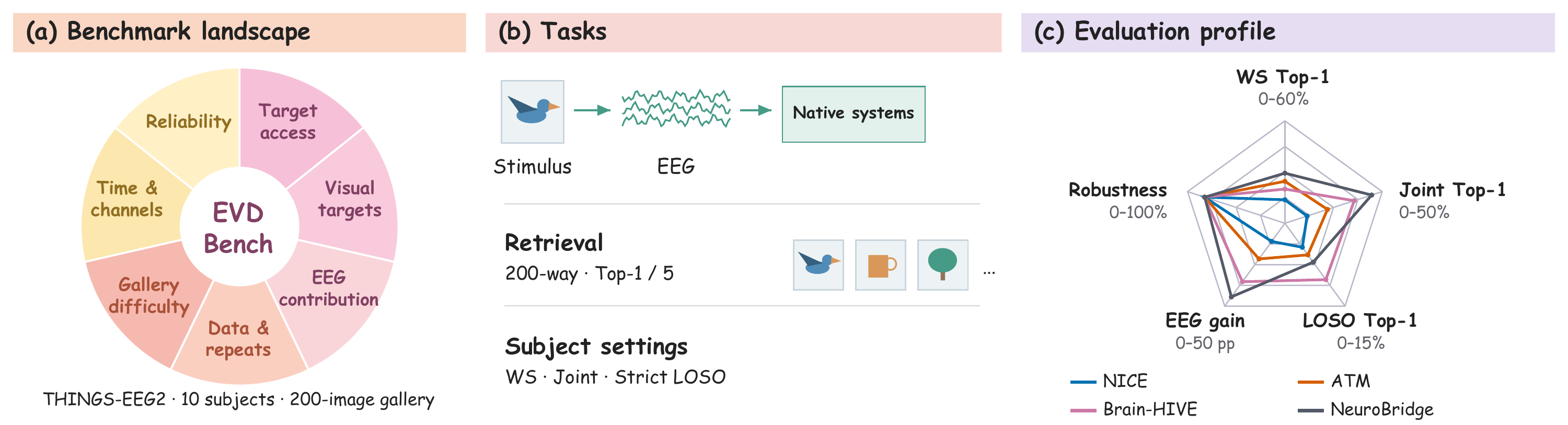}
\captionof{figure}{\textbf{We present EVD-Bench, a benchmark for EEG-to-image retrieval. (a) Seven evaluation topics on THINGS-EEG2 cover ten subjects and a shared 200-image test gallery. (b) Native systems rank the 200 images across three subject settings and are evaluated by Top-1 and Top-5. (c) Five axes show the measured profiles of four selected systems. EEG gain is the Joint Top-1 difference between correctly paired and matched-wrong EEG (pp), and robustness is WS accuracy retained at a reference SNR of 10\,dB (\% of clean-input Top-1).}}
\label{fig:evd-overview}
\label{fig:overview}
\end{minipage}
  \par\endgroup
  \vspace{8pt}%
}
\makeatother

\begin{document}
\twocolumn
\raggedbottom
\maketitle
\begingroup
\setlength{\parskip}{0pt}
\begin{abstract}
\normalfont\normalsize
\setlength{\baselineskip}{11.5pt}
EEG-to-image decoding recovers visual information from non-invasive
brain recordings, providing a window into human perception.
However, comparisons across systems are confounded by a variable that is
rarely reported as such: which participants' data the system was allowed
to use.
The same Top-1 score on a 200-way gallery may come from a per-participant
system, a pooled model over all participants, or a model that never saw
the test participant---cases that differ by more than most architectural
changes between the methods being compared.
We present \bench, a benchmark on THINGS-EEG2 that standardizes the gallery
and EEG input, matches training budgets across systems within each regime,
and compares three participant-access regimes: within-subject, joint-subject, and strict
leave-one-subject-out, with unlabeled target calibration reported
separately.
All three settings evaluate concept-level zero-shot retrieval: the 200 test
concepts are disjoint from the 1,654 concepts used for EEG--image fitting
and validation.
We compare retrieval across access regimes and use zero-EEG and
matched-wrong EEG controls to assess dependence on correct EEG--stimulus
pairing. We vary trial averaging, gallery size and difficulty, and
training data coverage, and compare target adaptation and visual feature
targets. We also assess sensitivity to EEG window length and channel
count, robustness to input perturbations, and confidence calibration.
These results establish participant-access regime as a necessary reporting dimension alongside gallery, repetition, and visual target.
\par 
\end{abstract}
\endgroup

\section{Introduction}
\label{sec:introduction}

Understanding how the brain encodes visual information is a central
question in cognitive neuroscience and a stepping stone toward
brain-computer interfaces~\citep{haxby2011common}.
Electroencephalography (EEG) offers a non-invasive, temporally precise
window into neural activity, but its low spatial resolution and high noise
have long made decoding visual content from EEG a challenging
task~\citep{spampinato2017eegvisual}.
The release of large-scale datasets like THINGS-EEG2~\citep{evbP01}---with
repeated EEG responses to thousands of natural images---has enabled
data-driven approaches to this problem.
A series of recent methods have demonstrated that retrieval of the viewed
image from a candidate set and reconstruction of its visual features are
now achievable with EEG alone~\citep{evbP02,evbP03,evbP05,evbP06,evbP07,
evbP09,evbP10,li2025neuralmcrl,rajabi2025aligned,liu2026visualblur,wu2026ats,
liu2025vieeg}.

Among these tasks, EEG-to-image retrieval---retrieving the viewed image
from a gallery based on single-trial or averaged EEG---has emerged as a
quantitative benchmark for evaluating decoding models.
State-of-the-art systems now achieve over 50\% Top-1 accuracy on 200-way
retrieval tasks, a striking improvement from near-chance performance just
a few years ago.
These advances span diverse architectures---contrastive
learning~\citep{evbP02,radford2021clip}, attention-based temporal
modeling~\citep{evbP03}, cross-modal alignment~\citep{evbP09}, and hybrid
approaches~\citep{evbP06,evbP07}---reflecting the field's rapid exploration
of model design.
Beyond scientific interest, such systems hold promise for assistive
communication devices and personalized content delivery based on neural
responses.

Yet as the field matures, a critical evaluation challenge has emerged:
reported accuracies are difficult to compare across papers.
A single paper can report two very different numbers for the same task.
NICE reports 13.8\% Top-1 when a system is trained on the participant it
is tested on, and 6.2\% when that participant is held out
\citep{evbP02}---same 200-image gallery, same 80-repeat test averaging,
same frozen image encoder.
This 2.2$\times$ difference illustrates the performance gap between
participant-specific training and transfer to an unseen participant.
This makes cross-paper numbers hard to interpret.
A reported 25\% Top-1 may come from a per-participant system trained on
that participant, from a pooled model over all ten participants, or from
a model that never saw the test participant at all---three cases that
differ by more than most architectural changes between the methods being
compared.
Several further conditions move with them: trial averaging (1--80 repeats),
the visual target (CLIP~\citep{radford2021clip}, DINOv2~\citep{oquab2024dinov2},
or ResNet-50~\citep{he2016resnet}), the candidate gallery, and---in at least one widely used
protocol---checkpoint selection on the test set itself \citep{evbP03}.
Without holding these fixed, a ranking between two systems conflates the
architecture with the access regime under which it happened to be
evaluated.

We therefore organize comparison around \emph{participant access}.
\textbf{\bench} evaluates EEG-to-image retrieval on THINGS-EEG2 under
three regimes defined by which participants' data the
system may use (Figure~\ref{fig:evd-overview}):
within-subject (WS) trains and tests per participant;
joint-subject trains one model on all ten and tests each;
and strict leave-one-subject-out (LOSO) trains and selects on nine
sources and tests the tenth.
All three draw queries from the same 200-image gallery, receive the same
standardized input, and use common epoch, batch, and selection rules.
Table~\ref{tab:retrieval} therefore compares participant-access settings
on a fixed retrieval task, with training data and total updates set by
the number of eligible participants.
Retrieval is concept-level zero-shot for EEG--image fitting: systems are
fitted on 1,488 concepts and selected on 166 validation concepts, then
rank the correct image among 200 images from 200 disjoint test concepts.

The same design lets us ask a second question that aggregate accuracy
cannot answer: how much of a system's score depends on the EEG actually
being paired with the image that was shown?
For each system we recompute retrieval under matched-wrong, shuffled,
and zero EEG, holding checkpoints, queries, and gallery fixed.
The resulting paired contrast measures dependence on the
stimulus--response pairing while keeping the system's visual components fixed.

We report all 13 systems across the three regimes, with ten-participant,
three-seed statistics in each setting.
Within-subject Top-1 spans 13.75\% to 58.65\%.
Joint training helps some systems substantially and hurts others.
Strict LOSO compresses all measured systems into a narrow range.
Paired matched-wrong controls show that retrieval depends heavily on
correct stimulus pairing under within-subject and joint access, but
less so for unseen participants.
These patterns establish baseline performance levels and demonstrate that
the three regimes are not interchangeable reporting conditions.

Our contributions are threefold.
\begin{itemize}[leftmargin=1em,labelsep=0.4em,itemindent=0pt,
  listparindent=0pt,topsep=2pt,itemsep=2pt,parsep=0pt,partopsep=0pt]
  \item A common protocol that makes WS, joint-subject, and strict LOSO
  directly comparable on one gallery, with data-access and
  selection rules stated per regime (Section~\ref{sec:benchmark}).
  \item Paired EEG--stimulus controls that quantify, per system and
  per regime, how much retrieval depends on correct pairing
  (Section~\ref{sec:eeg-dependence}).
  \item A 13-system evaluation along four axes---subject
  generalization, pairing dependence, repeat budget, and gallery
  difficulty
  (Sections~\ref{sec:result-tables}--\ref{sec:data-difficulty}).
\end{itemize}

\section{Related Work}
\label{sec:related}

\paragraph{EEG-Vision Decoding.}
Contrastive EEG--image alignment provides a route to zero-shot retrieval
of natural images excluded from EEG--image fitting, evaluated by
ranking the correct test image among unseen candidates, as exemplified by
NICE~\citep{evbP02}.
NICE reports both within-subject and held-out-subject experiments on
200-way THINGS-EEG2 retrieval with 80-repeat test averaging and
CLIP visual targets~\citep{evbP02,radford2021clip}.
ATM extends alignment to guided image reconstruction and evaluates both
within-subject and leave-one-subject-out regimes, though it reports the
highest accuracy observed during evaluation as its
statistic~\citep{evbP03}.
BrainFLORA uses multimodal neural embeddings for shared visual
representation learning~\citep{evbP06}.

Subsequent systems vary the visual supervision and EEG representation:
UBP introduces an uncertainty-aware blur prior to improve brain--vision
alignment~\citep{evbP05},
Neural-MCRL applies multimodal contrastive representation
learning~\citep{li2025neuralmcrl},
NeuroBridge combines cognitive priors with bidirectional semantic
alignment~\citep{evbP09},
and AlignVis studies human-aligned image encoders~\citep{rajabi2025aligned}.
Visual Blur changes the visual targets through blur~\citep{liu2026visualblur},
ATS adapts the visual teacher~\citep{wu2026ats}, and HyFI uses hyperbolic
feature interpolation~\citep{evbP10}. ViEEG and Brain-HIVE use
hierarchical visual representations~\citep{liu2025vieeg,evbP07}.
The iTransformer architecture originated in time-series
forecasting~\citep{liu2024itransformer} and is used as an EEG backbone
in SATTC~\citep{evbP08}.
EEG object classification predates these retrieval settings~\citep{spampinato2017eegvisual}.
Their alignment objectives draw on contrastive representation
learning~\citep{chen2020simclr,radford2021clip}.
The visual targets span CLIP variants~\citep{radford2021clip,cherti2023openclip},
DINOv2~\citep{oquab2024dinov2}, and task-specific
representations~\citep{liu2026visualblur}.

\paragraph{Evaluation Protocols and Benchmarks.}
Prior benchmarks address dimensions orthogonal to participant-access regime.
NeuralBench spans multiple neurophysiology tasks, including a
fixed-DINOv2 EEG image retrieval task~\citep{banville2026neuralbench,evbP14}.
EEG-EditBench evaluates subject-specific decoders on source--edit pairs
and same-source edit pools drawn from 2,137 controlled edits~\citep{evbP12}.
BASIC evaluates structural and semantic agreement at multiple
granularities~\citep{evbP11}.
CrossPT-EEG provides cross-participant and cross-time
evaluation~\citep{zhu2025crosspt}.
NECOMIMI studies image matching and generation within a shared
framework~\citep{evbP13}.
SATTC reports leave-one-subject-out evaluation with target adaptation,
showing substantial gains from standardized inference and unlabeled
target statistics~\citep{evbP08}.

Evaluation methodology in brain decoding draws on practices from computer
vision~\citep{geifman2017selective,guo2017calibration}, cross-modal
retrieval, and domain adaptation.
Retrieval uses ranking metrics such as Top-$k$ accuracy. Image
reconstruction is often evaluated with structural or perceptual
similarity measures~\citep{wang2004ssim,zhang2018lpips}.
Appendix~\ref{app:related-landscape} discusses the datasets, encoders,
and methodological precedents that inform these comparisons.

\section{Benchmark Construction}
\label{sec:benchmark}

\bench compares complete EEG-to-image systems on THINGS-EEG2 using
standardized EEG inputs, a shared retrieval task, three subject settings,
and explicit access, training-budget, and evaluation rules. Native visual
representations and objectives are retained. Paired controls measure
EEG-pairing dependence. Section~\ref{sec:experiments} reports comparisons;
Appendix~\ref{app:protocol} gives implementation details.

\subsection{Dataset and Input Standardization}
\label{sec:benchmark-data}
THINGS-EEG2 contains ten participants, each with 82,160 trials spanning
16,740 image conditions~\citep{evbP01}, drawn from the THINGS stimulus
collection~\citep{hebart2019things}. Its training partition comprises
1,654 concepts with ten images per concept and four presentations per
image; the test partition contains 200 separate concepts with one image
and 80 presentations per concept. We define validation by reserving
$\lceil0.1C_{\mathrm{train}}\rceil$ official training concepts in a fixed
hash order. Every image and trial of a concept stays in the same
partition, and all subjects share this split. Validation concepts are
not merged back into training, and test data cannot guide model selection.
The split contains 1,488 training and 166 validation concepts,
corresponding to 14,880 and 1,660 images per participant.

The common EEG input covers the onset-aligned interval $[0,1)$\,s at
200\,Hz, with the same ordered 63 EEG channels and no trigger channel.
Source timestamps define alignment. We crop 1,000 samples from stimulus
onset and resample from 1,000 to 200\,Hz using a polyphase filter,
without additional baseline correction or multivariate noise normalization.
Training and validation use four-repeat averages; testing uses 80-repeat
averages, computed in float64 and stored in float32. Channel-wise z-scoring
uses the permitted training pairs, with equal weight per source participant;
these statistics remain fixed at evaluation. Native model normalization
layers are retained. Appendix~\ref{app:protocol} specifies the input operator.

\subsection{Tasks and Subject-Generalization Protocols}
\label{sec:benchmark-tasks}
\begin{figure}[t]
\centering
\includegraphics[width=\linewidth]{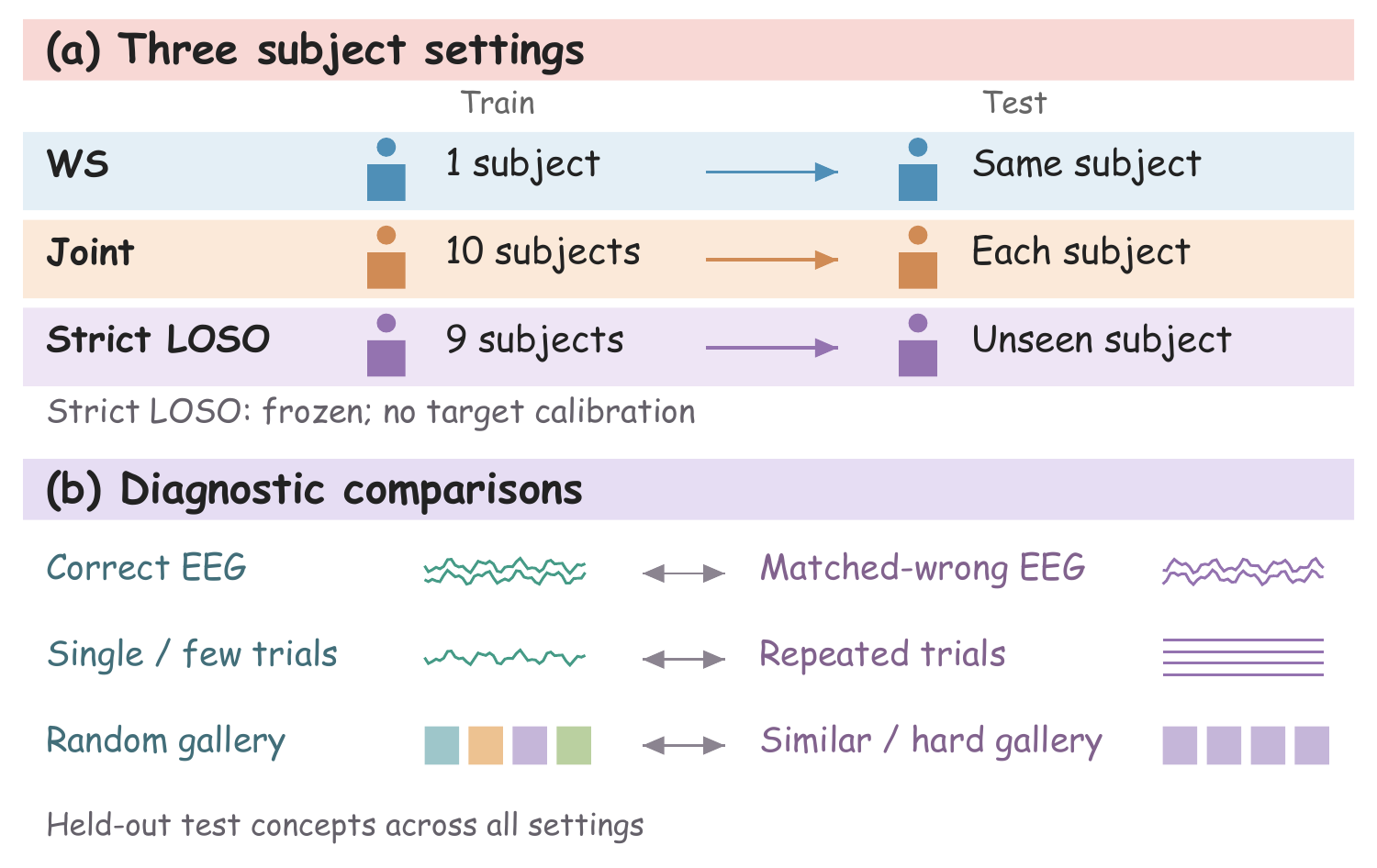}
\caption{\textbf{Subject settings and diagnostic comparisons.}
(a) Within-subject (WS) trains and tests on the same participant. Joint trains one model on all ten participants and tests each separately. Strict leave-one-subject-out (LOSO) trains on nine participants and tests the held-out participant without target calibration. All three settings use unseen test concepts.
(b) Diagnostics compare correct and matched-wrong EEG, averages of few and many test trials, and random and hard galleries of equal size.}
\label{fig:evd-settings}
\label{fig:settings}
\end{figure}

\begin{figure*}[t]
\centering
\includegraphics[width=\linewidth]{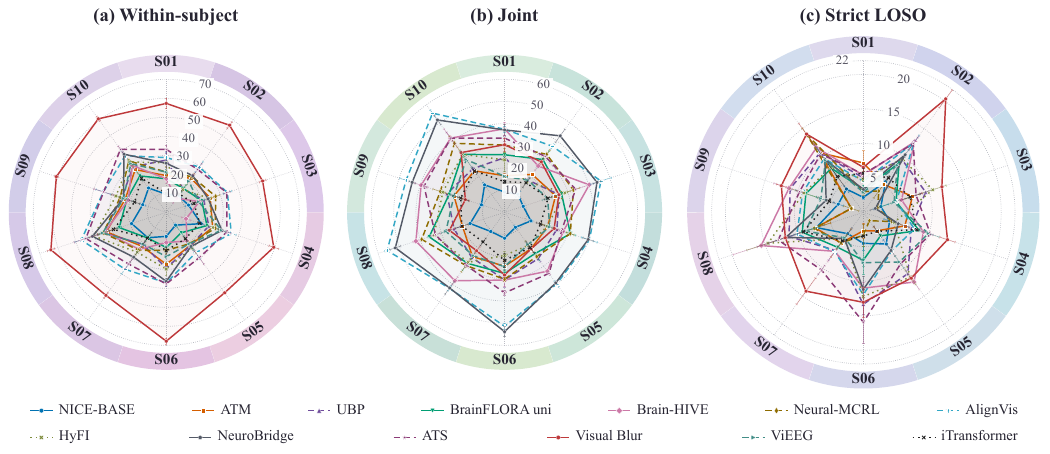}
\caption{\textbf{Subject-wise retrieval for 13 systems.}
Markers show mean Top-1 over three training seeds at fixed S01--S10 angles. Radial bars show $\pm$1 seed SD. Linear radial ranges are 0--70\% (WS), 0--60\% (Joint), and 0--22\% (strict LOSO). The LOSO panel is enlarged, so compare scores using the tick labels. Table~\ref{tab:retrieval} gives the exact values in columns S01--S10.}
\label{fig:subject-profiles}
\end{figure*}

For subject $s$ and stimulus $i$, let
$x_{s,i}\in\mathbb{R}^{63\times200}$ denote the EEG query after input
standardization. The retrieval gallery
$\mathcal{G}=\{I_j\}_{j=1}^{200}$ contains the actual test stimulus images,
with one ground-truth match $I_i$ per query. A system returns 200 finite
scores $f(x_{s,i},I_j)$; the evaluator obtains the positive index from the
stimulus manifest.
All subjects share the gallery, so 2,000 subject--query pairs represent
200 distinct stimuli.

All three settings are \emph{concept-level zero-shot} for EEG--image fitting: the 200 test
concepts are disjoint from the 1,654 training-partition concepts. No test image,
trial, concept label, or class prototype enters fitting or checkpoint
selection. Systems rank categories unseen in EEG--image fitting; subject novelty
is varied separately through strict LOSO. Visual pretraining may have included
these concepts, and its exposure is not assumed to be absent.

We define three training and evaluation settings plus a separate
calibration condition, illustrated in Figure~\ref{fig:evd-settings}(a).

\paragraph{Within-subject (WS).}
A separate system is trained on each participant's training partition
and evaluated on concepts unseen during EEG--image fitting from that same participant. The model
may fit participant-specific statistics, so the score describes a
personal decoder.

\paragraph{Joint-subject (Joint).}
One shared system is trained on all ten participants' training
partitions and evaluated separately on each participant's test queries.
The model sees every participant's labelled data but still reports a
per-participant score.

\paragraph{Strict leave-one-subject-out (LOSO).}
Nine source participants provide training data, and the held-out tenth
provides test queries, rotating across all ten folds. Every EEG-dependent
stage---encoder, visual teacher, subject embeddings, normalization, and
hyperparameter and checkpoint selection---uses source data only.
Inference uses frozen parameters and source statistics, with no
target-cohort fitting.

\paragraph{Target calibration.}
A fourth, separate condition relaxes only the last rule: model parameters
and source statistics stay frozen, but unlabeled target EEG may be used
to estimate normalization or similarity structure at deployment. We
distinguish an independent calibration cohort (I4) from calibration on
the test cohort itself (T4 with four trials, T80 with eighty), as in
SATTC~\citep{evbP08}. Calibration is label-free and reported separately
from strict LOSO.

For each system the EEG architecture, visual targets, loss, and key
components remain fixed across settings; routing for an unseen participant
is defined from source data before evaluation. Cross-setting contrasts
describe complete-system performance with the data quantities and update
counts associated with each access regime. Table~\ref{tab:retrieval}
reports ten-participant, three-seed statistics for all 13 systems in
each of the three settings.

\subsection{Data Access and Fair Comparison}
\label{sec:benchmark-access}
\begin{table*}[t]
\centering
\footnotesize
\setlength{\tabcolsep}{1pt}
\renewcommand{\arraystretch}{1.13}
\begin{tabular*}{\textwidth}{@{\extracolsep{\fill}}l*{11}{c}@{}}
\toprule
\multicolumn{12}{l}{\textbf{(a) Within-subject (WS)}} \\
Method & S01 & S02 & S03 & S04 & S05 & S06 & S07 & S08 & S09 & S10 & AVG \\
\midrule
NICE-BASE~\citep{evbP02} & \evdTableValue{9.50}{2.18} & \evdTableValue{10.83}{1.04} & \evdTableValue{12.33}{1.76} & \evdTableValue{19.33}{1.15} & \evdTableValue{8.33}{1.04} & \evdTableValue{12.67}{0.58} & \evdTableValue{16.50}{0.50} & \evdTableValue{19.17}{0.58} & \evdTableValue{13.00}{1.50} & \evdTableValue{15.83}{0.76} & \evdTableValue{13.75}{3.83} \\
ATM~\citep{evbP03} & \evdTableValue{19.33}{2.36} & \evdTableValue{22.17}{2.36} & \evdTableValue{24.33}{1.89} & \evdTableValue{25.67}{3.69} & \evdTableValue{21.50}{1.32} & \evdTableValue{27.33}{1.04} & \evdTableValue{22.00}{3.12} & \evdTableValue{33.17}{4.80} & \evdTableValue{22.67}{0.58} & \evdTableValue{27.83}{2.57} & \evdTableValue{24.60}{4.03} \\
UBP~\citep{evbP05} & \evdTableValue{27.50}{4.77} & \evdTableValue{15.17}{1.53} & \evdTableValue{20.83}{2.02} & \evdTableValue{23.50}{2.29} & \evdTableValue{21.67}{2.75} & \evdTableValue{29.50}{1.00} & \evdTableValue{20.83}{1.76} & \evdTableValue{34.83}{3.40} & \evdTableValue{24.67}{1.15} & \evdTableValue{29.67}{1.15} & \evdTableValue{24.82}{5.67} \\
BrainFLORA uni~\citep{evbP06} & \evdTableValue{17.67}{0.29} & \evdTableValue{15.50}{2.18} & \evdTableValue{19.67}{1.61} & \evdTableValue{22.50}{0.87} & \evdTableValue{13.33}{4.73} & \evdTableValue{18.50}{2.60} & \evdTableValue{16.67}{1.76} & \evdTableValue{26.00}{1.32} & \evdTableValue{21.83}{2.84} & \evdTableValue{23.67}{2.84} & \evdTableValue{19.53}{3.95} \\
Brain-HIVE~\citep{evbP07} & \evdTableValue{17.67}{2.32} & \evdTableValue{9.17}{3.25} & \evdTableValue{18.33}{1.88} & \evdTableValue{10.97}{2.53} & \evdTableValue{18.67}{2.32} & \evdTableValue{15.89}{1.99} & \evdTableValue{24.08}{1.88} & \evdTableValue{34.08}{2.47} & \evdTableValue{21.00}{1.73} & \evdTableValue{30.50}{2.88} & \evdTableValue{20.04}{7.83} \\
Neural-MCRL~\citep{li2025neuralmcrl} & \evdTableValue{21.17}{2.89} & \evdTableValue{23.17}{2.02} & \evdTableValue{27.33}{2.25} & \evdTableValue{27.17}{1.76} & \evdTableValue{20.50}{3.50} & \evdTableValue{20.00}{1.32} & \evdTableValue{24.67}{2.84} & \evdTableValue{34.83}{3.06} & \evdTableValue{25.50}{3.04} & \evdTableValue{33.50}{4.09} & \evdTableValue{25.78}{5.13} \\
AlignVis~\citep{rajabi2025aligned} & \evdTableValue{28.83}{2.93} & \evdTableValue{\underline{29.50}}{3.50} & \evdTableValue{\underline{33.83}}{2.36} & \evdTableValue{\underline{36.00}}{2.65} & \evdTableValue{26.33}{4.19} & \evdTableValue{37.17}{1.89} & \evdTableValue{\underline{37.00}}{3.12} & \evdTableValue{\underline{47.00}}{3.50} & \evdTableValue{\underline{34.83}}{4.07} & \evdTableValue{36.83}{3.51} & \evdTableValue{34.73}{5.78} \\
HyFI~\citep{evbP10} & \evdTableValue{26.00}{0.87} & \evdTableValue{14.83}{3.51} & \evdTableValue{23.83}{1.53} & \evdTableValue{25.83}{3.33} & \evdTableValue{23.17}{3.62} & \evdTableValue{29.83}{3.55} & \evdTableValue{26.00}{1.50} & \evdTableValue{40.17}{2.47} & \evdTableValue{27.83}{2.52} & \evdTableValue{32.83}{0.29} & \evdTableValue{27.03}{6.60} \\
NeuroBridge~\citep{evbP09} & \evdTableValue{25.83}{2.02} & \evdTableValue{23.83}{0.76} & \evdTableValue{23.17}{0.76} & \evdTableValue{30.83}{2.02} & \evdTableValue{22.17}{1.53} & \evdTableValue{36.33}{1.76} & \evdTableValue{29.17}{3.21} & \evdTableValue{40.83}{3.88} & \evdTableValue{23.67}{0.76} & \evdTableValue{37.83}{1.26} & \evdTableValue{29.37}{6.83} \\
ATS~\citep{wu2026ats} & \evdTableValue{\underline{33.17}}{2.25} & \evdTableValue{28.00}{3.77} & \evdTableValue{33.50}{1.73} & \evdTableValue{33.50}{2.29} & \evdTableValue{\underline{29.50}}{4.27} & \evdTableValue{\underline{37.67}}{1.04} & \evdTableValue{33.67}{3.25} & \evdTableValue{44.50}{6.50} & \evdTableValue{34.67}{5.39} & \evdTableValue{\underline{41.00}}{2.00} & \evdTableValue{\underline{34.92}}{4.97} \\
Visual Blur~\citep{liu2026visualblur} & \evdTableValue{\textbf{57.33}}{2.84} & \evdTableValue{\textbf{56.67}}{2.08} & \evdTableValue{\textbf{53.33}}{1.89} & \evdTableValue{\textbf{59.50}}{2.65} & \evdTableValue{\textbf{52.33}}{0.29} & \evdTableValue{\textbf{67.83}}{1.44} & \evdTableValue{\textbf{54.00}}{1.80} & \evdTableValue{\textbf{63.83}}{2.02} & \evdTableValue{\textbf{60.83}}{1.04} & \evdTableValue{\textbf{60.83}}{1.44} & \evdTableValue{\textbf{58.65}}{4.91} \\
ViEEG~\citep{liu2025vieeg} & \evdTableValue{21.17}{1.26} & \evdTableValue{20.17}{1.76} & \evdTableValue{18.00}{1.80} & \evdTableValue{30.50}{2.29} & \evdTableValue{20.50}{4.00} & \evdTableValue{23.33}{0.76} & \evdTableValue{20.67}{1.76} & \evdTableValue{35.17}{2.25} & \evdTableValue{23.67}{2.52} & \evdTableValue{31.67}{1.04} & \evdTableValue{24.48}{5.83} \\
iTransformer~\citep{evbP08} & \evdTableValue{10.00}{1.80} & \evdTableValue{11.83}{2.25} & \evdTableValue{15.33}{1.26} & \evdTableValue{18.50}{1.32} & \evdTableValue{17.17}{1.61} & \evdTableValue{20.17}{2.02} & \evdTableValue{16.83}{3.69} & \evdTableValue{29.17}{2.47} & \evdTableValue{17.50}{2.78} & \evdTableValue{22.17}{1.44} & \evdTableValue{17.87}{5.36} \\
\midrule
\multicolumn{12}{l}{\textbf{(b) Joint}} \\
Method & S01 & S02 & S03 & S04 & S05 & S06 & S07 & S08 & S09 & S10 & AVG \\
\midrule
NICE-BASE~\citep{evbP02} & \evdTableValue{9.17}{1.53} & \evdTableValue{10.17}{1.76} & \evdTableValue{8.83}{0.29} & \evdTableValue{12.50}{0.87} & \evdTableValue{8.33}{1.04} & \evdTableValue{11.67}{0.29} & \evdTableValue{10.67}{0.58} & \evdTableValue{16.33}{0.58} & \evdTableValue{11.00}{0.87} & \evdTableValue{15.33}{1.26} & \evdTableValue{11.40}{2.67} \\
ATM~\citep{evbP03} & \evdTableValue{16.50}{1.00} & \evdTableValue{21.17}{0.58} & \evdTableValue{24.17}{1.76} & \evdTableValue{20.50}{3.50} & \evdTableValue{20.33}{1.76} & \evdTableValue{24.33}{2.57} & \evdTableValue{24.67}{2.36} & \evdTableValue{20.00}{0.87} & \evdTableValue{24.00}{4.27} & \evdTableValue{23.17}{1.26} & \evdTableValue{21.88}{2.63} \\
UBP~\citep{evbP05} & \evdTableValue{24.67}{0.58} & \evdTableValue{18.00}{0.50} & \evdTableValue{25.83}{2.02} & \evdTableValue{24.83}{0.76} & \evdTableValue{25.83}{1.76} & \evdTableValue{30.50}{3.28} & \evdTableValue{25.17}{1.15} & \evdTableValue{24.00}{3.04} & \evdTableValue{27.33}{2.02} & \evdTableValue{24.33}{1.04} & \evdTableValue{25.05}{3.12} \\
BrainFLORA uni~\citep{evbP06} & \evdTableValue{25.98}{0.90} & \evdTableValue{29.45}{4.73} & \evdTableValue{27.03}{2.63} & \evdTableValue{31.22}{2.43} & \evdTableValue{23.11}{3.76} & \evdTableValue{27.56}{1.23} & \evdTableValue{28.45}{3.93} & \evdTableValue{35.77}{0.64} & \evdTableValue{26.29}{0.61} & \evdTableValue{32.33}{1.53} & \evdTableValue{28.72}{3.63} \\
Brain-HIVE~\citep{evbP07} & \evdTableValue{\textbf{37.67}}{2.98} & \evdTableValue{25.83}{2.40} & \evdTableValue{40.17}{6.11} & \evdTableValue{27.81}{0.82} & \evdTableValue{32.83}{0.38} & \evdTableValue{30.56}{3.02} & \evdTableValue{38.14}{2.15} & \evdTableValue{42.06}{1.47} & \evdTableValue{39.58}{3.19} & \evdTableValue{41.39}{5.15} & \evdTableValue{35.60}{5.89} \\
Neural-MCRL~\citep{li2025neuralmcrl} & \evdTableValue{30.33}{3.06} & \evdTableValue{32.33}{2.47} & \evdTableValue{32.67}{2.89} & \evdTableValue{31.17}{1.76} & \evdTableValue{24.67}{0.76} & \evdTableValue{30.17}{0.76} & \evdTableValue{29.67}{3.62} & \evdTableValue{39.33}{0.29} & \evdTableValue{30.50}{3.50} & \evdTableValue{38.50}{3.97} & \evdTableValue{31.93}{4.28} \\
AlignVis~\citep{rajabi2025aligned} & \evdTableValue{\underline{37.48}}{1.29} & \evdTableValue{\underline{36.83}}{1.16} & \evdTableValue{\textbf{45.68}}{4.92} & \evdTableValue{\textbf{40.24}}{3.74} & \evdTableValue{\textbf{39.97}}{1.70} & \evdTableValue{\underline{51.42}}{1.41} & \evdTableValue{\textbf{42.35}}{1.29} & \evdTableValue{\textbf{55.72}}{1.11} & \evdTableValue{\textbf{46.74}}{2.96} & \evdTableValue{\textbf{55.30}}{1.75} & \evdTableValue{\textbf{45.17}}{7.02} \\
HyFI~\citep{evbP10} & \evdTableValue{23.67}{5.25} & \evdTableValue{17.04}{0.51} & \evdTableValue{28.58}{2.56} & \evdTableValue{20.16}{2.36} & \evdTableValue{18.37}{2.51} & \evdTableValue{19.45}{1.74} & \evdTableValue{27.89}{1.86} & \evdTableValue{33.43}{0.94} & \evdTableValue{29.32}{1.05} & \evdTableValue{32.36}{1.42} & \evdTableValue{25.03}{6.04} \\
NeuroBridge~\citep{evbP09} & \evdTableValue{37.17}{2.75} & \evdTableValue{\textbf{42.67}}{3.62} & \evdTableValue{\underline{43.83}}{1.76} & \evdTableValue{\underline{39.67}}{2.47} & \evdTableValue{\underline{39.33}}{0.29} & \evdTableValue{\textbf{53.83}}{2.31} & \evdTableValue{\underline{42.17}}{3.55} & \evdTableValue{\underline{51.83}}{2.52} & \evdTableValue{\underline{43.83}}{1.53} & \evdTableValue{\underline{51.50}}{2.60} & \evdTableValue{\underline{44.58}}{5.81} \\
ATS~\citep{wu2026ats} & \evdTableValue{33.47}{1.34} & \evdTableValue{31.22}{0.75} & \evdTableValue{33.70}{0.27} & \evdTableValue{27.03}{0.90} & \evdTableValue{34.05}{0.82} & \evdTableValue{36.34}{1.04} & \evdTableValue{29.47}{0.96} & \evdTableValue{32.80}{2.21} & \evdTableValue{38.22}{1.35} & \evdTableValue{41.33}{3.22} & \evdTableValue{33.76}{4.15} \\
Visual Blur~\citep{liu2026visualblur} & \evdTableValue{30.50}{3.04} & \evdTableValue{28.15}{1.17} & \evdTableValue{25.74}{0.45} & \evdTableValue{23.52}{1.76} & \evdTableValue{20.38}{1.97} & \evdTableValue{28.25}{0.90} & \evdTableValue{21.43}{2.32} & \evdTableValue{26.18}{2.45} & \evdTableValue{18.60}{1.97} & \evdTableValue{33.33}{0.29} & \evdTableValue{25.61}{4.66} \\
ViEEG~\citep{liu2025vieeg} & \evdTableValue{16.24}{0.25} & \evdTableValue{17.92}{1.51} & \evdTableValue{20.42}{2.50} & \evdTableValue{21.38}{1.52} & \evdTableValue{17.97}{1.00} & \evdTableValue{24.52}{1.36} & \evdTableValue{24.12}{1.54} & \evdTableValue{27.83}{2.08} & \evdTableValue{24.03}{1.79} & \evdTableValue{29.44}{4.72} & \evdTableValue{22.39}{4.37} \\
iTransformer~\citep{evbP08} & \evdTableValue{13.82}{0.56} & \evdTableValue{16.50}{2.60} & \evdTableValue{20.61}{1.67} & \evdTableValue{17.06}{1.91} & \evdTableValue{19.65}{3.30} & \evdTableValue{21.70}{3.25} & \evdTableValue{16.51}{2.00} & \evdTableValue{20.25}{3.03} & \evdTableValue{20.82}{3.74} & \evdTableValue{23.09}{1.77} & \evdTableValue{19.00}{2.88} \\
\midrule
\multicolumn{12}{l}{\textbf{(c) Strict LOSO}} \\
Method & S01 & S02 & S03 & S04 & S05 & S06 & S07 & S08 & S09 & S10 & AVG \\
\midrule
NICE-BASE~\citep{evbP02} & \evdTableValue{2.17}{0.29} & \evdTableValue{5.50}{0.87} & \evdTableValue{2.67}{1.15} & \evdTableValue{4.00}{0.87} & \evdTableValue{5.67}{0.76} & \evdTableValue{4.50}{1.00} & \evdTableValue{3.83}{0.76} & \evdTableValue{7.17}{1.26} & \evdTableValue{3.83}{0.76} & \evdTableValue{4.17}{0.58} & \evdTableValue{4.35}{1.46} \\
ATM~\citep{evbP03} & \evdTableValue{\textbf{7.17}}{1.76} & \evdTableValue{5.00}{1.32} & \evdTableValue{7.50}{1.50} & \evdTableValue{6.33}{0.76} & \evdTableValue{3.33}{0.76} & \evdTableValue{2.67}{1.15} & \evdTableValue{5.50}{2.50} & \evdTableValue{7.33}{2.36} & \evdTableValue{1.83}{0.29} & \evdTableValue{10.67}{1.61} & \evdTableValue{5.73}{2.65} \\
UBP~\citep{evbP05} & \evdTableValue{5.00}{0.50} & \evdTableValue{9.50}{0.87} & \evdTableValue{9.50}{0.87} & \evdTableValue{7.33}{1.04} & \evdTableValue{11.00}{0.50} & \evdTableValue{\underline{13.33}}{1.53} & \evdTableValue{6.83}{1.76} & \evdTableValue{11.33}{0.76} & \evdTableValue{11.17}{0.29} & \evdTableValue{10.67}{1.89} & \evdTableValue{9.57}{2.50} \\
BrainFLORA uni~\citep{evbP06} & \evdTableValue{3.33}{0.58} & \evdTableValue{8.83}{0.29} & \evdTableValue{5.00}{2.18} & \evdTableValue{8.50}{0.87} & \evdTableValue{4.22}{0.26} & \evdTableValue{6.95}{1.97} & \evdTableValue{6.29}{0.81} & \evdTableValue{9.31}{2.63} & \evdTableValue{8.76}{1.61} & \evdTableValue{7.90}{1.73} & \evdTableValue{6.91}{2.12} \\
Brain-HIVE~\citep{evbP07} & \evdTableValue{5.39}{0.47} & \evdTableValue{\underline{12.67}}{1.04} & \evdTableValue{5.83}{0.58} & \evdTableValue{9.17}{0.72} & \evdTableValue{\textbf{12.50}}{1.32} & \evdTableValue{11.00}{0.87} & \evdTableValue{7.00}{1.80} & \evdTableValue{\textbf{15.58}}{4.25} & \evdTableValue{\underline{11.33}}{2.32} & \evdTableValue{11.58}{0.52} & \evdTableValue{10.21}{3.30} \\
Neural-MCRL~\citep{li2025neuralmcrl} & \evdTableValue{3.00}{0.50} & \evdTableValue{3.83}{1.26} & \evdTableValue{2.17}{0.29} & \evdTableValue{4.33}{0.29} & \evdTableValue{1.50}{0.00} & \evdTableValue{3.33}{1.61} & \evdTableValue{5.50}{1.50} & \evdTableValue{7.50}{2.50} & \evdTableValue{1.83}{1.15} & \evdTableValue{\underline{13.83}}{1.26} & \evdTableValue{4.68}{3.69} \\
AlignVis~\citep{rajabi2025aligned} & \evdTableValue{4.51}{0.09} & \evdTableValue{12.45}{1.34} & \evdTableValue{5.04}{0.15} & \evdTableValue{7.78}{0.26} & \evdTableValue{6.97}{0.05} & \evdTableValue{11.94}{0.05} & \evdTableValue{7.09}{0.22} & \evdTableValue{10.12}{3.21} & \evdTableValue{9.26}{0.35} & \evdTableValue{9.64}{1.61} & \evdTableValue{8.48}{2.68} \\
HyFI~\citep{evbP10} & \evdTableValue{4.78}{0.30} & \evdTableValue{9.62}{0.38} & \evdTableValue{\underline{10.58}}{2.06} & \evdTableValue{4.71}{0.40} & \evdTableValue{11.60}{1.34} & \evdTableValue{12.22}{0.24} & \evdTableValue{4.67}{0.32} & \evdTableValue{\underline{14.54}}{3.22} & \evdTableValue{9.07}{2.15} & \evdTableValue{8.06}{0.19} & \evdTableValue{8.99}{3.44} \\
NeuroBridge~\citep{evbP09} & \evdTableValue{3.33}{0.29} & \evdTableValue{10.33}{2.02} & \evdTableValue{1.83}{0.76} & \evdTableValue{5.33}{1.61} & \evdTableValue{6.83}{1.04} & \evdTableValue{11.17}{1.89} & \evdTableValue{4.67}{1.89} & \evdTableValue{11.83}{2.57} & \evdTableValue{5.67}{1.26} & \evdTableValue{9.50}{2.65} & \evdTableValue{7.05}{3.47} \\
ATS~\citep{wu2026ats} & \evdTableValue{5.52}{0.18} & \evdTableValue{12.33}{1.09} & \evdTableValue{8.04}{0.89} & \evdTableValue{\underline{10.25}}{0.38} & \evdTableValue{9.22}{0.24} & \evdTableValue{\textbf{15.94}}{3.01} & \evdTableValue{\underline{10.34}}{0.80} & \evdTableValue{12.98}{3.22} & \evdTableValue{9.73}{0.46} & \evdTableValue{10.50}{1.40} & \evdTableValue{\underline{10.49}}{2.83} \\
Visual Blur~\citep{liu2026visualblur} & \evdTableValue{\underline{6.53}}{0.58} & \evdTableValue{\textbf{20.34}}{1.59} & \evdTableValue{\textbf{11.98}}{0.02} & \evdTableValue{\textbf{12.84}}{1.16} & \evdTableValue{\underline{11.81}}{0.27} & \evdTableValue{13.05}{2.24} & \evdTableValue{\textbf{14.14}}{2.34} & \evdTableValue{11.90}{0.27} & \evdTableValue{\textbf{12.50}}{0.73} & \evdTableValue{\textbf{14.03}}{1.90} & \evdTableValue{\textbf{12.91}}{3.37} \\
ViEEG~\citep{liu2025vieeg} & \evdTableValue{2.91}{0.08} & \evdTableValue{8.44}{0.06} & \evdTableValue{4.68}{0.44} & \evdTableValue{9.94}{1.60} & \evdTableValue{8.91}{1.08} & \evdTableValue{7.30}{0.55} & \evdTableValue{4.56}{0.26} & \evdTableValue{8.83}{1.49} & \evdTableValue{6.47}{0.72} & \evdTableValue{9.34}{2.21} & \evdTableValue{7.14}{2.39} \\
iTransformer~\citep{evbP08} & \evdTableValue{5.67}{0.76} & \evdTableValue{6.18}{0.33} & \evdTableValue{7.14}{1.56} & \evdTableValue{8.14}{0.28} & \evdTableValue{3.39}{0.83} & \evdTableValue{3.45}{0.08} & \evdTableValue{5.11}{0.08} & \evdTableValue{9.14}{0.14} & \evdTableValue{5.19}{0.14} & \evdTableValue{6.43}{2.24} & \evdTableValue{5.99}{1.85} \\
\bottomrule
\end{tabular*}
\caption{\textbf{Subject-wise retrieval across three settings.} Top-1 accuracy (\%) on the shared 200-image gallery. S01--S10 report mean $\pm$ seed SD over three training seeds. AVG reports the equally weighted mean $\pm$ SD across the ten subject means. In each column within a setting, including AVG, bold marks the highest mean and underlining marks the second-highest.}
\label{tab:retrieval}
\end{table*}

Within each setting, methods receive the same eligible EEG--image pairs,
repeat budgets, data-exposure limits, and validation and tuning
opportunities. The comparison retains native visual representations,
objectives, and system components. All methods use 200 epochs and batch
size 128; the number of source participants determines total exposure and
updates (Section~\ref{sec:setup}). This is a comparison of complete
systems; model size and GPU time are not matching criteria.

Strict LOSO applies the source-only rule to the entire EEG-dependent
chain: encoder, teacher, paired visual alignment, EEG prior, subject
embeddings, normalization, and hyperparameter and checkpoint selection.
A Joint parent model exposed to target EEG cannot initialize a strict
LOSO run. Fixed visual pretraining is permitted with its sources and
possible test-image exposure disclosed. Test-image overlap in visual
pretraining is unknown when its training data cannot be audited.
External EEG/MEG/fMRI-paired supervision and
unknown brain-subject exposure require separate reporting, and unresolved
eligibility excludes a system from the EEG-only primary comparison.

Target normalization, batch-statistic updates, whitening, and cross-query
calibration constitute unlabeled target adaptation even when model
weights are frozen, as illustrated by SATTC~\citep{evbP08}.
Independent-target and test-cohort adaptation are reported separately
from strict LOSO; labeled target adaptation is outside the current
comparison. Strict LOSO predictions must be invariant to query batching
and ordering. Independent query and gallery permutations test whether
outputs depend on pairing information implicit in their order.

\subsection{Paired Controls for EEG Contribution}
\label{sec:benchmark-controls}
\begin{figure*}[t]
\centering
\includegraphics[width=\linewidth]{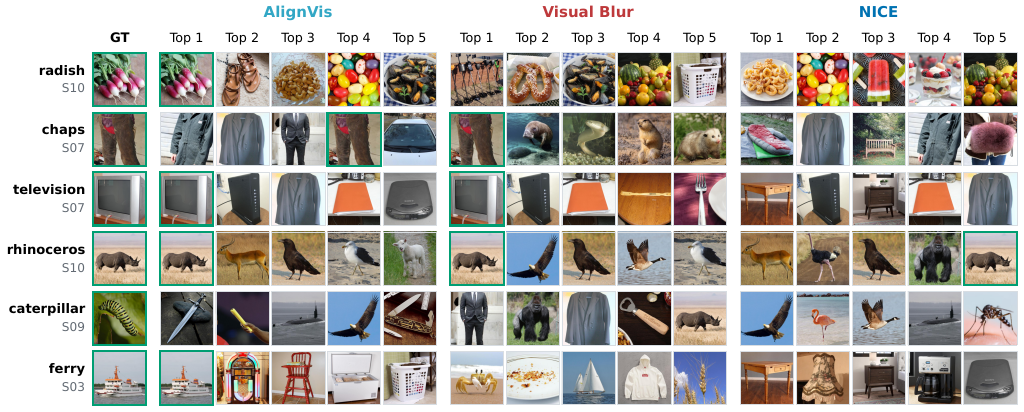}
\caption{\textbf{Joint retrieval examples.}
Each row shows one subject--query pair. Six distinct concepts are sampled from 2,000 pairs across ten subjects without conditioning on retrieval results. The left column shows ground truth (GT). AlignVis, Visual Blur, and NICE are ranked first, seventh, and thirteenth by mean Joint Top-1 among 13 methods. The top five images use archived seed-0 scores. Green borders mark GT wherever it appears.}
\label{fig:retrieval-examples}
\end{figure*}

\begin{figure}[!tb]
\centering
\includegraphics[width=\linewidth]{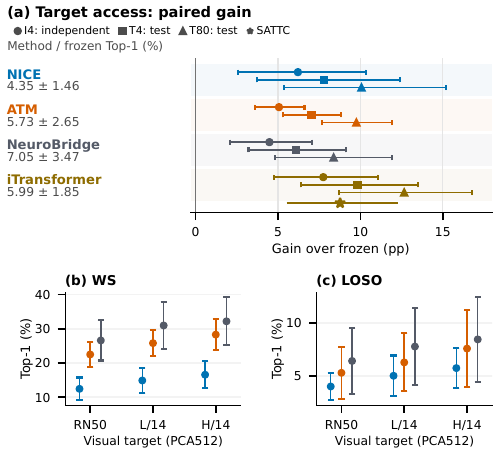}
\caption{\textbf{Target access and visual configuration.}
(a) Left labels give frozen strict-LOSO Top-1 (\%, mean $\pm$ subject SD). Points show subject-paired gains over frozen inference (pp) with descriptive 95\% intervals. Unlabeled calibration uses four repeats per stimulus in an independent cohort (I4) or four or 80 repeats in the test cohort (T4/T80). Only iTransformer uses SATTC, with T80 input.
(b,c) Points and bars give Top-1 mean $\pm$ subject SD for RN50, L/14, and H/14 under WS and strict LOSO. Colors follow (a). The 512-dimensional PCA is fitted on training images only.}
\label{fig:access-visual-eeg}
\end{figure}

\begin{table*}[t]
\centering
\footnotesize
\setlength{\tabcolsep}{3pt}
\renewcommand{\arraystretch}{1.0}
\begin{tabular*}{\textwidth}{@{\extracolsep{\fill}}lcccccc@{}}
\toprule
& \multicolumn{2}{c}{Within-subject} & \multicolumn{2}{c}{Joint} & \multicolumn{2}{c}{Strict LOSO} \\
\cmidrule(lr){2-3}\cmidrule(lr){4-5}\cmidrule(lr){6-7}
Method & $\Delta$Zero & $\Delta$Matched & $\Delta$Zero & $\Delta$Matched & $\Delta$Zero & $\Delta$Matched \\
\midrule
NICE-BASE~\citep{evbP02} & \evdTableValue{13.25}{3.83} & \evdTableValue{13.28}{3.75} & \evdTableValue{10.90}{2.67} & \evdTableValue{10.97}{2.78} & \evdTableValue{3.85}{1.46} & \evdTableValue{4.03}{1.45} \\
ATM~\citep{evbP03} & \evdTableValue{24.10}{4.03} & \evdTableValue{24.20}{3.93} & \evdTableValue{21.38}{2.63} & \evdTableValue{21.53}{2.57} & \evdTableValue{5.23}{2.65} & \evdTableValue{5.37}{2.69} \\
UBP~\citep{evbP05} & \evdTableValue{24.32}{5.67} & \evdTableValue{24.53}{5.65} & \evdTableValue{24.55}{3.12} & \evdTableValue{24.65}{3.29} & \evdTableValue{9.07}{2.50} & \evdTableValue{9.23}{2.62} \\
BrainFLORA uni~\citep{evbP06} & \evdTableValue{19.03}{3.95} & \evdTableValue{19.18}{3.98} & \evdTableValue{28.22}{3.63} & \evdTableValue{28.27}{3.61} & \evdTableValue{6.41}{2.12} & \evdTableValue{6.53}{2.11} \\
Brain-HIVE~\citep{evbP07} & \evdTableValue{19.54}{7.83} & \evdTableValue{19.76}{7.81} & \evdTableValue{35.10}{5.89} & \evdTableValue{35.29}{5.90} & \evdTableValue{9.71}{3.30} & \evdTableValue{9.92}{3.52} \\
Neural-MCRL~\citep{li2025neuralmcrl} & \evdTableValue{25.28}{5.13} & \evdTableValue{25.55}{5.18} & \evdTableValue{31.43}{4.28} & \evdTableValue{31.82}{4.27} & \evdTableValue{4.18}{3.69} & \evdTableValue{4.15}{3.70} \\
AlignVis~\citep{rajabi2025aligned} & \evdTableValue{34.23}{5.78} & \evdTableValue{34.47}{5.79} & \evdTableValue{\textbf{44.67}}{7.02} & \evdTableValue{\textbf{44.97}}{7.02} & \evdTableValue{7.98}{2.68} & \evdTableValue{7.91}{2.88} \\
HyFI~\citep{evbP10} & \evdTableValue{26.53}{6.60} & \evdTableValue{26.78}{6.44} & \evdTableValue{24.53}{6.04} & \evdTableValue{24.61}{5.97} & \evdTableValue{8.49}{3.44} & \evdTableValue{8.47}{3.25} \\
NeuroBridge~\citep{evbP09} & \evdTableValue{28.87}{6.83} & \evdTableValue{29.03}{6.88} & \evdTableValue{\underline{44.08}}{5.81} & \evdTableValue{\underline{44.47}}{5.86} & \evdTableValue{6.55}{3.47} & \evdTableValue{6.83}{3.55} \\
ATS~\citep{wu2026ats} & \evdTableValue{\underline{34.42}}{4.97} & \evdTableValue{\underline{34.73}}{4.94} & \evdTableValue{33.26}{4.15} & \evdTableValue{33.35}{4.28} & \evdTableValue{\underline{9.99}}{2.83} & \evdTableValue{\underline{10.05}}{2.78} \\
Visual Blur~\citep{liu2026visualblur} & \evdTableValue{\textbf{58.15}}{4.91} & \evdTableValue{\textbf{58.42}}{4.78} & \evdTableValue{25.11}{4.66} & \evdTableValue{25.36}{4.70} & \evdTableValue{\textbf{12.41}}{3.37} & \evdTableValue{\textbf{12.39}}{3.34} \\
ViEEG~\citep{liu2025vieeg} & \evdTableValue{23.98}{5.83} & \evdTableValue{24.15}{5.84} & \evdTableValue{21.89}{4.37} & \evdTableValue{22.14}{4.35} & \evdTableValue{6.64}{2.39} & \evdTableValue{6.64}{2.47} \\
iTransformer~\citep{evbP08} & \evdTableValue{17.37}{5.36} & \evdTableValue{17.42}{5.25} & \evdTableValue{18.50}{2.88} & \evdTableValue{18.70}{3.01} & \evdTableValue{5.49}{1.85} & \evdTableValue{5.59}{1.87} \\
\bottomrule
\end{tabular*}
\caption{\textbf{Paired Top-1 gains over EEG controls.}
Correct-minus-zero ($\Delta$Zero) and correct-minus-matched-wrong ($\Delta$Matched) gains are in percentage points. Each entry gives mean $\pm$ SD across ten subjects after averaging three training seeds per subject. Both controls share the checkpoint, queries, and gallery with the correct condition. Matched-wrong EEG comes from a different stimulus with the same participant, session composition, and repeat count. Zero EEG is an out-of-distribution diagnostic. In each column, bold marks the highest mean gain and underlining marks the second-highest. Gains use the original paired runs.}
\label{tab:eeg-pairing}
\end{table*}

Paired controls measure how task performance depends on correctly
matched EEG (Figure~\ref{fig:evd-settings}(b)). The primary control substitutes EEG from a wrong stimulus
while matching subject, session composition, and repeat count.
Subject-wise shuffled EEG and zero input provide
additional diagnostics. Matched-wrong and shuffled controls provide
complementary evidence when their mappings differ. Zero input is an
out-of-distribution diagnostic.

Within a system, controls share the checkpoint, evaluation queries,
and gallery. Every EEG-derived condition is recomputed from the
substituted input. For metric $m$, the primary
paired gain is
\begin{equation}
 \Delta_m=d_m\bigl(m_{\mathrm{correct}}-m_{\mathrm{matched}}\bigr),
 \label{eq:eeg-gain}
\end{equation}
where $d_m=+1$ for retrieval accuracy.
Thus positive gain favors correct EEG.
The gain measures dependence on correct EEG pairing for the specified
task and gallery.

\subsection{Evaluation Metrics and Statistical Reporting}
\label{sec:benchmark-evaluation}
\paragraph{Task metrics.}
Retrieval uses Top-1 and Top-5 with uniform tie breaking. If $a_i$
candidates score strictly above the positive and its tied group contains
$b_i$ candidates including the positive, its Top-$k$ contribution is
\begin{equation}
 h_k(i)=\min\!\left(1,\max\!\left(0,\frac{k-a_i}{b_i}\right)\right).
 \label{eq:topk}
\end{equation}
Accuracy is $100\,\mathrm{mean}_i\,h_k(i)$, with chance levels of
0.5\% and 2.5\% for this 200-way single-positive task.
\paragraph{Aggregation and uncertainty.}
Queries are first averaged within each subject--seed unit,
then the three seeds within subjects,
and finally the ten subjects with equal weight. Table~\ref{tab:retrieval} reports each subject's mean and seed SD.
Aggregate reporting includes subject
SD ($\mathrm{ddof}=1$) and 95\% percentile intervals from 10,000
subject-bootstrap resamples (seed 42)~\citep{efron1979bootstrap}.
Contrasts resample subject-paired
differences with common indices. Accuracy gains use percentage points;
other gains retain the metric's native units. Effect sizes and paired
intervals are primary.

\section{Experiment and Analysis}
\label{sec:experiments}

We examine retrieval performance across three participant-access regimes,
paired EEG-stimulus controls, and the effects of trial averaging and
gallery size under Section~\ref{sec:benchmark}'s rules.

\subsection{Experimental Setup}
\label{sec:setup}
\paragraph{Systems and training.}
We compare NICE-BASE~\citep{evbP02}, ATM~\citep{evbP03}, UBP~\citep{evbP05},
BrainFLORA uni~\citep{evbP06}, Brain-HIVE~\citep{evbP07},
Neural-MCRL~\citep{li2025neuralmcrl}, AlignVis~\citep{rajabi2025aligned},
HyFI~\citep{evbP10}, NeuroBridge~\citep{evbP09}, ATS~\citep{wu2026ats},
Visual Blur~\citep{liu2026visualblur}, ViEEG~\citep{liu2025vieeg}, and
SATTC's uncalibrated iTransformer backbone~\citep{evbP08}.
BrainFLORA uses its EEG-only uni route; iTransformer excludes target
calibration; NeuroBridge uses TSConv in all settings. Chance and deterministic ridge fits
serve as separate references.

Training uses seeds $\{0,1,2\}$, 200 epochs, batch size 128, and no
early stopping. Author optimizers and learning rates are retained without
search. Each epoch visits every eligible EEG--image pair once and retains
the final partial batch. WS, Joint, and strict LOSO use 23,400, 234,000,
and 210,600 updates, respectively. Shared-model sampling balances subjects
and excludes same-stimulus cross-subject negatives. Validation runs every
epoch over all 1,660 validation images. Checkpoints maximize
source-subject-macro Top-1, with ties resolved toward earlier epochs.

\subsection{Retrieval Under Three Access Regimes}
\label{sec:result-tables}
\label{sec:retrieval-analysis}

Table~\ref{tab:retrieval} reports mean $\pm$ seed SD over three seeds
for each setting, system, and subject. AVG gives the equally weighted
mean and SD across the ten subject means.
Figure~\ref{fig:subject-profiles} shows the per-subject profiles. Appendix
Tables~\ref{tab:s3}--\ref{tab:s5} retain detailed summaries.

\paragraph{Within-subject and Joint performance.}
\begin{figure}[tb]
\centering
\includegraphics[width=\linewidth]{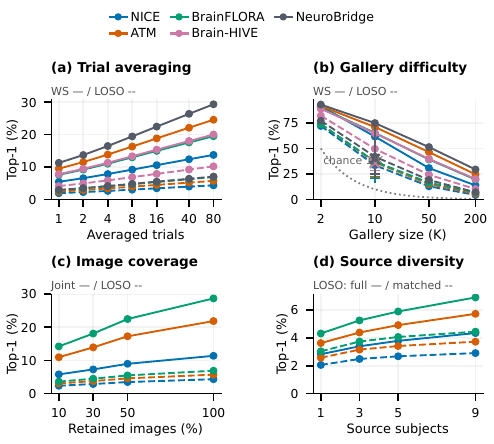}
\caption{\textbf{Data requirements.}
Curves show subject-macro Top-1 means (\%).
(a) Test-trial averaging with fixed checkpoints.
(b) Candidate count $K$, with chance at $100/K$ (dotted gray). At $K=10$, vertical lines connect random and hard galleries. Hard-gallery results use $\times$ for WS and $+$ for LOSO. Solid and dashed curves in (a,b) denote WS and LOSO.
(c) Training images retained per concept (solid Joint, dashed LOSO). Nominal 25\% retains three of ten images, so it is plotted at 30\%.
(d) Number of LOSO source subjects, using all available data (solid) or matching sample and update counts (dashed). Panels (c,d) show NICE-BASE, ATM, and BrainFLORA uni.}
\label{fig:data-requirements}
\end{figure}

Visual Blur leads WS Top-1 at 58.65\%, followed by ATS (34.92\%) and
AlignVis (34.73\%). Joint training benefits several systems:
NeuroBridge reaches 44.58\% versus 29.37\% under WS, Brain-HIVE
35.60\% versus 20.04\%, and Neural-MCRL 31.93\% versus 25.78\%.
UBP changes little (25.05\% versus 24.82\%), while NICE-BASE and ATM
decrease by 2.35pp and 2.72pp.

\paragraph{Strict LOSO and qualitative retrieval.}
LOSO performance drops across systems: Visual Blur reaches 12.91\%,
ATS 10.49\%, Brain-HIVE 10.21\%, and UBP 9.57\%.
The WS--LOSO gaps include 9.40pp for NICE-BASE and 22.32pp for
NeuroBridge, with substantial individual differences in
Figure~\ref{fig:subject-profiles}.
Figure~\ref{fig:retrieval-examples} shows AlignVis, Visual Blur, and
NICE, ranked best, median, and worst (1, 7, 13) by mean Joint Top-1
across all 13 methods. The examples include hits, misses, and
query-level rank reversals. Appendix~\ref{app:qualitative-selection}
specifies the sampling rule and gives further examples.

\subsection{EEG Pairing Contribution}
\label{sec:eeg-dependence}

Table~\ref{tab:eeg-pairing} reports correct-minus-zero and
correct-minus-matched-wrong Top-1 gains for all 13 systems, as mean
$\pm$ subject SD after averaging seeds. Matched-wrong input substitutes
another stimulus's EEG while preserving participant and session conditions.
NeuroBridge's matched gain is $44.47\pm5.86$pp under Joint and
$6.83\pm3.55$pp under LOSO. Neural-MCRL achieves 4.68\% LOSO
Top-1, with a matched-control gain of $4.15\pm3.70$pp.

\subsection{Target Adaptation and Visual Configuration}
\label{sec:target-adaptation}

Figure~\ref{fig:access-visual-eeg}(a) separates frozen source-only
LOSO from independent-cohort (I4) and test-cohort (T4/T80) adaptation.
For NICE-BASE, ATM, and NeuroBridge, T80 gives the largest reported
gain: Top-1 rises from 4.35\% to 14.42\%, 5.73\% to 15.50\%, and
7.05\% to 15.43\%, respectively. NeuroBridge reaches 11.53\% with
I4 and 13.16\% with T4; even its T80 result remains below Joint
performance (44.58\%). Panels (b,c) compare RN50,
L/14, and H/14 visual targets through the common train-only PCA512 interface.

\subsection{Trial Averaging and Gallery Size}
\label{sec:data-difficulty}

Figure~\ref{fig:data-requirements}(a) shows increasing accuracy with
trial averaging. From one to 80 repeats, WS Top-1 increases from
9.51\% to 24.60\% for ATM and from 7.80\% to 20.04\% for Brain-HIVE;
NICE-BASE reaches 13.75\% from 5.45\%. LOSO remains lower: ATM
rises from 2.45\% to 5.73\%, and Brain-HIVE from 4.13\% to 10.21\%.
Thus averaging helps but does not close the subject-generalization gap.
Panels (b--d) separate gallery size and difficulty, image coverage,
and source-subject diversity. The nominal 25\% image condition uses
three of ten images per concept (30\% actual retention).

\subsection{Predictive Reliability and Calibration}
\label{sec:reliability}
\begin{figure}[tb]
\centering
\includegraphics[width=\linewidth]{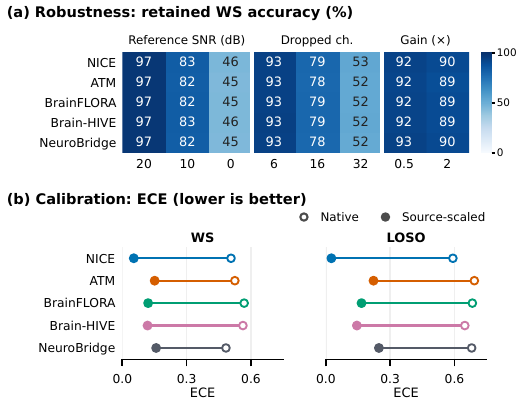}
\caption{\textbf{Robustness and calibration.}
(a) Heatmap values give the ratio of perturbed to clean WS subject-macro Top-1, in percent. At 100\%, accuracy matches the clean baseline. Noise, channel dropout, and gain are applied after 80-trial averaging.
(b) Colors identify methods. Open and filled markers give native and temperature-scaled expected calibration error (ECE, lower is better) under WS and LOSO. Scaling uses temperatures fitted on source validation data and preserves within-query ranks. Figure~\ref{fig:access-visual-eeg}(a) describes calibration with unlabeled target EEG.}
\label{fig:sensitivity-reliability}
\end{figure}

Figure~\ref{fig:sensitivity-reliability} shows retained accuracy under
input perturbations and expected calibration error (ECE).
Temperature scaling reduces ECE substantially: for NICE-BASE, WS ECE
drops from 0.51 to 0.05 and LOSO ECE from 0.59 to 0.03.
Lower ECE need not improve selective prediction; ATM WS AURC
increases from 0.6894 to 0.6990 after scaling.
Source-only temperature scaling preserves within-query ranks.

\paragraph{Scope.}
The main retrieval and paired-control tables cover 13 systems, ten
participants, and three seeds in each setting on THINGS-EEG2.
Diagnostic analyses use the method subsets identified in their figures.

\section{Conclusion}
\label{sec:conclusion}

We introduced \bench to make EEG-to-image retrieval results comparable
across participant-access regimes through standardized inputs, explicit
training budgets, and a shared evaluation protocol.
Our results show that participant access, EEG--stimulus pairing, trial
averaging, and gallery conditions shape reported performance, with a
persistent generalization gap for unseen participants.
The current benchmark covers ten participants from THINGS-EEG2 in
controlled offline retrieval; broader datasets, cross-session tests, and
prospective single-trial evaluation are natural extensions.
By making these dependencies explicit, \bench can guide reproducible
model development toward reliable and accessible visual brain--computer
interfaces.

\clearpage
\phantomsection
\pdfbookmark[0]{References}{references}
{%
  \small
  \nocite{*}
  \bibliographystyle{ieeenat_fullname}
  \bibliography{main}
}

\clearpage
\appendix
\makeatletter
\newcommand{\SupItemAnchor}[1]{\Hy@raisedlink{\hyper@anchorstart{supp.#1}\hyper@anchorend}}
\newcommand{\SupItemLabel}[1]{\def\@currentHref{supp.#1}\label{#1}}
\newcommand{\SupTableAlias}[2]{\begingroup\def\@currentHref{supp.#1}\def\@currentlabel{\thetable}\label{#2}\endgroup}
\makeatother

\setcounter{table}{0}
\setcounter{figure}{0}
\renewcommand{\thetable}{S\arabic{table}}
\renewcommand{\thefigure}{S\arabic{figure}}
\renewcommand{\theHtable}{supp.\arabic{table}}
\renewcommand{\theHfigure}{supp.\arabic{figure}}
\definecolor{suppink}{HTML}{254963}
\definecolor{supprule}{HTML}{C8D5DC}
\newcommand{\D}{\textcolor{suppink}{\ensuremath{\bullet}}}
\onecolumn
\setlength{\multicolsep}{5pt plus 2pt minus 1pt}
\setlength{\premulticols}{32pt}
\setlength{\postmulticols}{12pt}
\setlength{\maxbalancingoverflow}{0pt}
\newenvironment{supplementwide}
  {\par\addvspace{4pt}\noindent\begin{minipage}{\linewidth}}
  {\par\vspace{6pt}\end{minipage}\par\addvspace{4pt}}
\newcommand{\SupAsset}[3]{\par\SupItemAnchor{#3}%
  \captionsetup{type=figure,hypcap=false,position=bottom,skip=4pt,labelfont=bf,justification=raggedright,singlelinecheck=false}%
  {\centering#1\par}%
  \caption{#2}\SupItemLabel{#3}\par}
\setlength{\emergencystretch}{1em}
\newcommand{\SupTableStyle}[1]{%
  \centering\fontsize{#1}{\dimexpr#1pt+2pt\relax}\selectfont
  \setlength{\tabcolsep}{3pt}\renewcommand{\arraystretch}{1.07}%
  \setlength{\heavyrulewidth}{.65pt}\setlength{\lightrulewidth}{.4pt}%
  \setlength{\cmidrulewidth}{.35pt}}
\newcommand{\SupCaption}[3]{\par\begingroup
  \SupItemAnchor{#3}\captionsetup{type=#1,hypcap=false,position=top,skip=4pt,labelfont=bf,justification=raggedright,singlelinecheck=false}\caption{#2}\SupItemLabel{#3}\endgroup\par\nobreak}
\raggedbottom
\setcounter{topnumber}{4}
\renewcommand{\topfraction}{0.88}
\renewcommand{\textfraction}{0.12}
\setlength{\textfloatsep}{8pt plus 2pt minus 2pt}
\setlength{\floatsep}{8pt plus 2pt minus 2pt}
\pdfbookmark[0]{Appendix}{appendix}
\begin{center}
{\Large\bfseries Appendix}
\end{center}
\begin{multicols}{2}
\section{Benchmark Scope and Related Work}
\subsection{Test Gallery and Subject Access}\label{app:related-landscape}
Ten THINGS-EEG2 participants each viewed 200 fixed THINGS photographs 80 times~\citep{evbP01,hebart2019things}. Averaging the 16,000 presentations per participant gives 200 retrieval decisions. Each test concept has one photograph, so these decisions cannot distinguish category recognition from discrimination between exemplars of the same concept. Within-subject (WS) training uses the evaluated participant, Joint uses all ten, and strict leave-one-subject-out (LOSO) uses the other nine.

EEG-EditBench uses edited versions of a source image~\citep{evbP12}. Our hard gallery selects visually similar images of different concepts. NeuralBench fixes its visual teacher~\citep{banville2026neuralbench}. Table~\ref{tab:systemimpl} lists the EEG and vision encoders used here. Section~\ref{app:targets} compares three visual teachers with the EEG output dimension held fixed.
\subsection{System Selection and Diagnostic Coverage}
Table~\ref{tab:coverage} lists the systems in each comparison. Native SATTC uses the iTransformer backbone and the target test cohort at the T80 budget. Brain-HIVE's native fusion is evaluated separately in Section~\ref{app:fusion}.
\begin{supplementwide}
\SupCaption{table}{\textbf{Diagnostic coverage by retrieval system.} Dots identify the systems evaluated in each diagnostic. All 13 systems have retrieval scores in all three access regimes. Section~\ref{app:confidence} identifies the subset with verified probabilities. Stress includes averaging, gallery difficulty and perturbations. Data covers image and source counts. Mask covers time and channel restrictions. Target denotes visual targets. Gen.\ denotes Joint reconstruction.}{tab:coverage}
\begingroup\SupTableStyle{7.6}\setlength{\tabcolsep}{1.6pt}
\begin{tabular*}{\linewidth}{@{\extracolsep{\fill}}l*{6}{c}@{}}\toprule
\textbf{System} & \textbf{Adapt.} & \textbf{Target} & \textbf{Stress} & \textbf{Data} & \textbf{Mask} & \textbf{Gen.}\\\midrule
NICE-BASE~\citep{evbP02} & \D & \D & \D & \D & \D & --\\
ATM~\citep{evbP03} & \D & \D & \D & \D & \D & \D\\
UBP~\citep{evbP05} & -- & -- & -- & -- & -- & --\\
BrainFLORA uni~\citep{evbP06} & -- & -- & \D & \D & -- & \D\\
Brain-HIVE~\citep{evbP07} & -- & -- & \D & -- & -- & \D\\
Neural-MCRL~\citep{li2025neuralmcrl} & -- & -- & -- & -- & -- & --\\
AlignVis~\citep{rajabi2025aligned} & -- & -- & -- & -- & -- & --\\
HyFI~\citep{evbP10} & -- & -- & -- & -- & -- & --\\
NeuroBridge~\citep{evbP09} & \D & \D & \D & -- & \D & --\\
ATS~\citep{wu2026ats} & -- & -- & -- & -- & -- & --\\
Visual Blur~\citep{liu2026visualblur} & -- & -- & -- & -- & -- & --\\
ViEEG~\citep{liu2025vieeg} & -- & -- & -- & -- & -- & --\\
iTransformer~\citep{liu2024itransformer,evbP08} & \D & -- & -- & -- & -- & --\\
\bottomrule\end{tabular*}\par\endgroup
\end{supplementwide}

\section{Benchmark Implementation}\label{app:protocol}
\begin{table*}[t]
\SupCaption{table}{\textbf{Encoders and training objectives for the 13 retrieval systems.} The vision target is the representation paired with the EEG during training. Brain-HIVE's encoders are those identified in its main Joint run (Section~\ref{app:fusion}).}{tab:systemimpl}
\begingroup\SupTableStyle{8.5}
\begin{tabularx}{\linewidth}{@{}>{\raggedright\arraybackslash}p{.16\linewidth}>{\raggedright\arraybackslash}p{.18\linewidth}>{\raggedright\arraybackslash}p{.15\linewidth}>{\hsize=.94\hsize\linewidth=\hsize\raggedright\arraybackslash}X>{\hsize=1.06\hsize\linewidth=\hsize\raggedright\arraybackslash}X@{}}\toprule
\textbf{System} & \textbf{EEG Encoder} & \textbf{Vision Encoder} & \textbf{Vision Target} & \textbf{Training Objective}\\\midrule
NICE-BASE~\citep{evbP02} & Temporal--spatial convolution and projection & CLIP ViT-L/14 & Frozen image embedding & Bidirectional EEG--image contrastive loss\\\addlinespace[2pt]
ATM~\citep{evbP03} & ATMS attention and convolutional projector & ViT-H/14 & Image and text embeddings & Contrastive EEG alignment with image and text embeddings\\\addlinespace[2pt]
UBP~\citep{evbP05} & EEGProjectLayer & RN50 & FoveaBlur image embeddings & Bidirectional contrastive loss with uncertainty-guided blur\\
BrainFLORA uni~\citep{evbP06} & EEG-only Medformer~\citep{wang2024medformer} & ViT-H/14 & Image and text embeddings & Weighted EEG--image and EEG--text contrastive losses\\\addlinespace[2pt]
Brain-HIVE~\citep{evbP07} & BrainMLP & SynCLR ViT-B/16~\citep{tian2024synclr}, CLIP ViT-B/32, VAE & Fused image embeddings and VAE latents & BrainCLIP contrastive loss on the fused representation\\\addlinespace[2pt]
Neural-MCRL~\citep{li2025neuralmcrl} & Subject-specific layer and iTransformer & ViT-H/14 & Image and text embeddings & Weighted EEG--image and EEG--text contrastive losses\\\addlinespace[2pt]
AlignVis~\citep{rajabi2025aligned} & NICE BrainEncoder & DreamSim ensemble~\citep{fu2023dreamsim} & Human-aligned image embedding (1,792 dimensions) & Bidirectional contrastive loss on normalized embeddings\\\addlinespace[2pt]
HyFI~\citep{evbP10} & Hyperbolic EEGProjectLayer & RN50 & Geodesic interpolation of semantic and perceptual image features & Hyperbolic contrastive loss\\\addlinespace[2pt]
NeuroBridge~\citep{evbP09} & TSConv and projectors & RN50 & Image embedding fused with a cognitive prior & Bidirectional EEG--image contrastive loss\\\addlinespace[2pt]
ATS~\citep{wu2026ats} & Shared temporal attention encoder & RN50 with shrink adapter & Adapted image embedding & Contrastive alignment with similarity, distribution-matching and mapping terms\\\addlinespace[2pt]
Visual Blur~\citep{liu2026visualblur} & Temporal--spatial Brain Visual Encoder & CLIP RN50 & Learned fusion of image features across blur levels & Bidirectional EEG--image contrastive loss\\\addlinespace[2pt]
ViEEG~\citep{liu2025vieeg} & Three-branch spatiotemporal encoder with cross-attention & ViT-H/14 & Concatenated image and mask features & EEG-to-image cross-entropy on concatenated branch embeddings\\\addlinespace[2pt]
iTransformer~\citep{liu2024itransformer,evbP08} & SATTC iTransformer backbone & ViT-H/14 & Image and text embeddings with source-fitted visual whitening & Weighted EEG--image and EEG--text contrastive losses\\
\bottomrule\end{tabularx}\par\endgroup
\end{table*}
\subsection{Data Splits and EEG Input Construction}
We use the same concept split for all ten THINGS-EEG2 participants~\citep{evbP01}. Of the 1,654 official training concepts, 1,488 are used for training and 166 for validation. All ten images of a concept and every EEG repeat of those images stay in the same partition. The validation gallery contains 1,660 images. The official test gallery contains one photograph from each of 200 separate concepts. The supplementary data list the concept IDs in each partition.

\paragraph{From raw samples to a query.}
We extracted one-second epochs aligned to stimulus onset, resampled them from 1,000 to 200 Hz, and retained 63 EEG channels. We averaged repeated presentations separately for each participant and stimulus. The supplementary data specify the channel order and resampling filter.

\paragraph{Normalization.}
For channel $c$, let $(\mu_{sc},v_{sc})$ be the population moments across training pairs and 200 time points of eligible source $s$. For $S$ such sources, compute
\[
 \begin{aligned}
 \mu_c&=S^{-1}\sum_s\mu_{sc},\\
 v_c&=S^{-1}\sum_s\{v_{sc}+(\mu_{sc}-\mu_c)^2\},\\
 x'_{ct}&=(x_{ct}-\mu_c)/\sqrt{v_c}.
 \end{aligned}
\]
We standardized each channel using training-set moments pooled with equal weight across source participants. The variance includes differences between participant means. The same transform applies at every test time point.

\subsection{Subject Access and Training Budgets}
Within-subject (WS) training uses the evaluated participant's labeled EEG and their own normalization and validation data. Joint training includes all ten participants, including the evaluated participant. Under strict LOSO, training, normalization, and model selection use only the other nine participants. Joint and strict LOSO weight eligible participants equally for normalization and validation. Training and validation use four repeats per image, and test queries average 80. The validation and test galleries contain 1,660 and 200 images, respectively.

\paragraph{Updates and checkpoint selection.}
Each epoch visits every training pair once in participant-specific batches of up to 128, including the final partial batch. WS uses one training participant per fit, with 14,880 EEG--image pairs and 23,400 updates over 200 epochs. Joint uses ten participants, 148,800 pairs and 234,000 updates per fit. Strict LOSO uses nine training participants, 133,920 pairs and 210,600 updates per fit. WS and strict LOSO each require ten fits per seed, while Joint requires one. For each of three seeds, we selected the earliest checkpoint with the highest participant-macro validation Top-1.

\paragraph{Held-out inference.}
We fitted normalization, subject routing and visual-teacher adaptation on eligible sources and froze the selected decoder for inference. Under strict LOSO, a query's prediction must be invariant to the order and batching of other test queries.

\begin{table*}[t]
\SupCaption{table}{\textbf{Training compute and model size across 13 retrieval systems.} Parameter counts are in millions (M). The accompanying CSV retains four decimal places.}{tab:training-compute}
\begingroup\SupTableStyle{8.8}
\begin{tabular*}{\linewidth}{@{\extracolsep{\fill}}lrrrrr@{}}\toprule
\multicolumn{1}{c}{\textbf{}} & \multicolumn{3}{c}{\textbf{Training compute (GPU-hours)}} & \multicolumn{2}{c}{\textbf{Parameters (M)}}\\
\cmidrule(lr){2-4}\cmidrule(lr){5-6}
\textbf{System} & \textbf{WS} & \textbf{Joint} & \textbf{Strict LOSO} & \textbf{Trainable} & \textbf{Frozen}\\\midrule
NICE-BASE & 5.77 & 5.44 & 49.65 & 2.68 & 303.97\\
ATM & 8.86 & 8.37 & 78.89 & 3.57 & 986.11\\
UBP & 12.32 & 26.26 & 217.27 & 13.96 & 38.32\\
BrainFLORA uni & 25.62 & 37.24 & 226.21 & 19.17 & 986.11\\
Brain-HIVE & 6.40 & 6.05 & 56.16 & 19.46 & 207.81\\
Neural-MCRL & 11.27 & 11.35 & 98.43 & 5.11 & 986.11\\
AlignVis & 25.32 & 24.14 & 224.20 & 8.40 & 259.95\\
HyFI & 16.62 & 15.91 & 146.68 & 16.06 & 38.32\\
NeuroBridge & 6.94 & 6.96 & 63.25 & 3.80 & 38.32\\
ATS & 65.90 & 63.79 & 599.75 & 10.73 & 38.32\\
Visual Blur & 51.49 & 49.65 & 459.15 & 6.83 & 38.32\\
ViEEG & 34.77 & 32.99 & 279.52 & 6.68 & 852.25\\
iTransformer backbone & 46.50 & 45.08 & 317.08 & 3.57 & 986.11\\
\bottomrule\end{tabular*}\par\endgroup

\end{table*}

\subsection{System-Specific Implementations and Reference Baselines}
Table~\ref{tab:systemimpl} gives the native configuration for each system. Section~\ref{app:targets} compares vision encoders while holding the EEG architecture and output dimension fixed.

\paragraph{Linear and chance references.}
The linear ridge baseline was evaluated on an earlier split. On the 200-image test gallery, uniform random ranking gives expected Top-1 and Top-5 accuracies of 0.5\% and 2.5\%.

\subsection{Retrieval Evaluation, Paired Controls, and Statistical Inference}
We matched each query to its target among the 200 candidates by stimulus ID. If $a$ candidates score strictly above the positive and $b$ share its score including itself, average over the positive's possible ranks $a+1,\ldots,a+b$. Its Top-$k$ contribution is $\min(1,\max(0,(k-a)/b))$. Its reciprocal-rank contribution is $b^{-1}\sum_{r=a+1}^{a+b}1/r$. If all candidates tie, the rule recovers uniform random expectations. Qualitative displays require a single ordering and break ties by candidate ID.

\paragraph{Pairing controls.}
We held the participant, queries, checkpoint, gallery and evaluator fixed across pairing controls. For shuffled EEG, we permuted the 200 stimuli within each participant, excluding self-matches (seed 17). For matched-wrong EEG, we selected a different stimulus from the same participant with the same session composition and repeat count (seed 23). We recomputed all EEG-dependent quantities after substitution. For the zero-input control, we replaced the EEG waveform with zeros before encoding. Correct minus matched-wrong is the primary contrast because both conditions contain a measured EEG response with the same acquisition budget.
Because all 80-repeat queries have the same session composition, both controls sample from the same set of incorrect stimuli.

\paragraph{Statistical unit.}
We averaged the 200 queries within each participant and seed, then averaged seeds 0, 1 and 2 within participant. We report the arithmetic mean and sample SD across ten participants. For paired intervals, we computed the difference within each participant before resampling. We use a participant-level percentile bootstrap~\citep{efron1979bootstrap}. The 95\% intervals use the 2.5th and 97.5th percentiles of 10,000 subject-bootstrap draws (seed 42). Within each seed the ten Joint evaluations share a checkpoint. Two LOSO folds share eight training participants. Subject resampling describes variation in this cohort, conditional on the fitted models and the fixed gallery.

\subsection{Training Compute and Model Size}\label{app:training-compute}
Table~\ref{tab:training-compute} reports the total training GPU-hours for each system under WS, Joint and strict LOSO. Parameter counts are reported per model, with shared weights counted once. Frozen counts include the offline encoders used to construct cached targets.

\section{Detailed Retrieval Results}\label{app:retrieval}
\subsection{Retrieval Across Three Access Regimes}
Tables~\ref{tab:retrieval-ws}, \ref{tab:retrieval-joint} and \ref{tab:retrieval-loso} give Top-5 and mean reciprocal rank (MRR) across the ten participants for WS, Joint and strict LOSO. We averaged the three seeds within each participant before calculating the mean and sample SD across participants. Under Joint training, AlignVis and NeuroBridge have similar Top-5 accuracy (77.80\% and 77.78\%) and MRR (0.5940 and 0.5902). Their Top-1 ordering reverses between seeds 0 and 2. Participant-level results, seed-level summaries and paired intervals are included in the supplementary data.

\begin{table*}[t]
\SupCaption{table}{\textbf{Within-subject retrieval.} Values are participant means $\pm$ SD.}{tab:retrieval-ws}
\SupTableAlias{tab:retrieval-ws}{tab:s3}
\begingroup\SupTableStyle{8.8}
\begin{tabular*}{\linewidth}{@{\extracolsep{\fill}}lrrlrr@{}}\toprule
\textbf{System} & \textbf{Top-5 (\%)} & \textbf{MRR} & \textbf{System} & \textbf{Top-5 (\%)} & \textbf{MRR}\\\midrule
NICE-BASE & 39.07 \ensuremath{\pm} 8.47 & 0.2657 \ensuremath{\pm} 0.0536 & HyFI & 58.97 \ensuremath{\pm} 8.68 & 0.4209 \ensuremath{\pm} 0.0700\\
ATM & 57.03 \ensuremath{\pm} 6.69 & 0.3957 \ensuremath{\pm} 0.0454 & NeuroBridge & 61.75 \ensuremath{\pm} 8.46 & 0.4434 \ensuremath{\pm} 0.0711\\
UBP & 55.08 \ensuremath{\pm} 6.59 & 0.3903 \ensuremath{\pm} 0.0593 & ATS & 67.22 \ensuremath{\pm} 5.59 & 0.4958 \ensuremath{\pm} 0.0467\\
BrainFLORA uni & 47.82 \ensuremath{\pm} 6.92 & 0.3343 \ensuremath{\pm} 0.0481 & Visual Blur & 87.55 \ensuremath{\pm} 3.33 & 0.7125 \ensuremath{\pm} 0.0392\\
Brain-HIVE & 48.54 \ensuremath{\pm} 13.03 & 0.3383 \ensuremath{\pm} 0.0936 & ViEEG & 55.57 \ensuremath{\pm} 7.22 & 0.3927 \ensuremath{\pm} 0.0605\\
Neural-MCRL & 56.90 \ensuremath{\pm} 6.23 & 0.4028 \ensuremath{\pm} 0.0529 & iTransformer & 47.72 \ensuremath{\pm} 7.41 & 0.3223 \ensuremath{\pm} 0.0584\\
AlignVis & 68.92 \ensuremath{\pm} 5.99 & 0.5018 \ensuremath{\pm} 0.0570 &  &  & \\
\bottomrule\end{tabular*}\par\endgroup

\smallskip
\SupCaption{table}{\textbf{Joint retrieval.} Values are participant means $\pm$ SD.}{tab:retrieval-joint}
\begingroup\SupTableStyle{8.8}
\begin{tabular*}{\linewidth}{@{\extracolsep{\fill}}lrrlrr@{}}\toprule
\textbf{System} & \textbf{Top-5 (\%)} & \textbf{MRR} & \textbf{System} & \textbf{Top-5 (\%)} & \textbf{MRR}\\\midrule
NICE-BASE & 35.42 \ensuremath{\pm} 5.59 & 0.2392 \ensuremath{\pm} 0.0374 & HyFI & 55.58 \ensuremath{\pm} 9.00 & 0.3963 \ensuremath{\pm} 0.0663\\
ATM & 53.52 \ensuremath{\pm} 4.76 & 0.3671 \ensuremath{\pm} 0.0262 & NeuroBridge & 77.78 \ensuremath{\pm} 5.57 & 0.5902 \ensuremath{\pm} 0.0555\\
UBP & 54.88 \ensuremath{\pm} 5.50 & 0.3907 \ensuremath{\pm} 0.0357 & ATS & 63.88 \ensuremath{\pm} 4.57 & 0.4771 \ensuremath{\pm} 0.0395\\
BrainFLORA uni & 60.86 \ensuremath{\pm} 3.83 & 0.4349 \ensuremath{\pm} 0.0343 & Visual Blur & 57.06 \ensuremath{\pm} 4.82 & 0.4069 \ensuremath{\pm} 0.0447\\
Brain-HIVE & 68.96 \ensuremath{\pm} 7.42 & 0.5062 \ensuremath{\pm} 0.0608 & ViEEG & 51.78 \ensuremath{\pm} 6.57 & 0.3636 \ensuremath{\pm} 0.0485\\
Neural-MCRL & 65.16 \ensuremath{\pm} 4.95 & 0.4683 \ensuremath{\pm} 0.0413 & iTransformer & 46.94 \ensuremath{\pm} 2.72 & 0.3286 \ensuremath{\pm} 0.0271\\
AlignVis & 77.80 \ensuremath{\pm} 5.36 & 0.5940 \ensuremath{\pm} 0.0603 &  &  & \\
\bottomrule\end{tabular*}\par\endgroup

\smallskip
\SupCaption{table}{\textbf{Strict LOSO retrieval.} Values are participant means $\pm$ SD.}{tab:retrieval-loso}
\SupTableAlias{tab:retrieval-loso}{tab:s5}
\begingroup\SupTableStyle{8.8}
\begin{tabular*}{\linewidth}{@{\extracolsep{\fill}}lrrlrr@{}}\toprule
\textbf{System} & \textbf{Top-5 (\%)} & \textbf{MRR} & \textbf{System} & \textbf{Top-5 (\%)} & \textbf{MRR}\\\midrule
NICE-BASE & 15.80 \ensuremath{\pm} 3.51 & 0.1163 \ensuremath{\pm} 0.0233 & HyFI & 26.39 \ensuremath{\pm} 6.45 & 0.1862 \ensuremath{\pm} 0.0442\\
ATM & 18.13 \ensuremath{\pm} 6.31 & 0.1348 \ensuremath{\pm} 0.0429 & NeuroBridge & 23.82 \ensuremath{\pm} 8.79 & 0.1640 \ensuremath{\pm} 0.0553\\
UBP & 26.12 \ensuremath{\pm} 6.41 & 0.1919 \ensuremath{\pm} 0.0408 & ATS & 27.03 \ensuremath{\pm} 5.14 & 0.2030 \ensuremath{\pm} 0.0368\\
BrainFLORA uni & 20.00 \ensuremath{\pm} 4.83 & 0.1468 \ensuremath{\pm} 0.0329 & Visual Blur & 33.87 \ensuremath{\pm} 5.52 & 0.2430 \ensuremath{\pm} 0.0404\\
Brain-HIVE & 29.17 \ensuremath{\pm} 7.73 & 0.2085 \ensuremath{\pm} 0.0488 & ViEEG & 21.71 \ensuremath{\pm} 5.28 & 0.1567 \ensuremath{\pm} 0.0371\\
Neural-MCRL & 15.15 \ensuremath{\pm} 8.23 & 0.1145 \ensuremath{\pm} 0.0554 & iTransformer & 18.89 \ensuremath{\pm} 4.12 & 0.1406 \ensuremath{\pm} 0.0286\\
AlignVis & 25.80 \ensuremath{\pm} 8.24 & 0.1798 \ensuremath{\pm} 0.0440 &  &  & \\
\bottomrule\end{tabular*}\par\endgroup

\end{table*}

\subsection{Qualitative Retrieval Examples and Sampling Procedure}\label{app:qualitative-selection}

We sampled subject--image pairs in a fixed random order and retained the first occurrence of each concept. Selection used only participant and image identifiers. Main Figure~\ref{fig:retrieval-examples} shows items 1--6 and Figure~\ref{fig:supp-continuation} shows items 7--12 of the same draw.

Figures~\ref{fig:supp-examples}--\ref{fig:supp-ranks-11-13} use a second draw spanning ten participants and 30 distinct concepts. Each plate shows three concepts for each of two participants. The first model group spans ranks 1, 7 and 13 by three-seed Joint Top-1. The other four groups cover ranks 2--13 in order. The supplementary data list the selected query IDs, predictions and ranks.

\subsection{Paired Effects of Subject Access and EEG--Stimulus Pairing}\label{app:pairing}
Two pairings answer different questions. Comparing access regimes pairs the same participant across different trained models. Comparing inputs pairs the same checkpoint and gallery with different EEG assignments. Figure~\ref{fig:access-pairs} gives the former comparison. Visual Blur loses 33.04 percentage points (pp) from WS to Joint, whereas Brain-HIVE gains 15.57 pp. Each contrast changes both the training participants and the total number of updates. Neither contrast alone estimates the effect of adding a participant at a fixed budget.
\begin{supplementwide}
\SupAsset{\includegraphics[width=\linewidth]{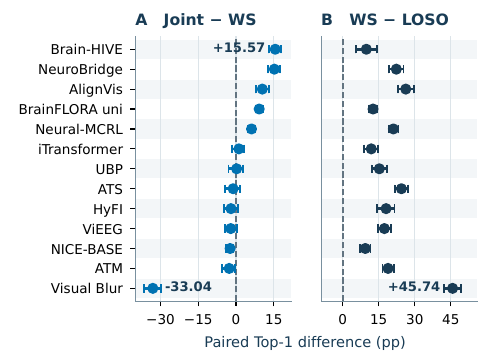}}{\textbf{Top-1 differences between subject-access regimes.} Points and bars show paired means and 95\% bootstrap intervals. Both panels follow Joint-minus-WS order.}{fig:access-pairs}
\end{supplementwide}

\begin{table*}[t]
\SupCaption{table}{\textbf{Top-1 accuracy (\%) with shuffled and matched-wrong EEG.} Zero EEG gives 0.50\% in all 39 system--regime cells. Main Table~\ref{tab:retrieval} gives correct EEG.}{tab:abs-control}
\begingroup\SupTableStyle{8.8}
\begin{tabular*}{\linewidth}{@{\extracolsep{\fill}}lrrrrrr@{}}\toprule
\multicolumn{1}{c}{\textbf{}} & \multicolumn{2}{c}{\textbf{WS}} & \multicolumn{2}{c}{\textbf{Joint}} & \multicolumn{2}{c}{\textbf{Strict LOSO}}\\
\cmidrule(lr){2-3}\cmidrule(lr){4-5}\cmidrule(lr){6-7}
\textbf{System} & \textbf{Shuffled} & \textbf{Matched-wrong} & \textbf{Shuffled} & \textbf{Matched-wrong} & \textbf{Shuffled} & \textbf{Matched-wrong}\\\midrule
NICE-BASE & 0.18 & 0.47 & 0.20 & 0.43 & 0.48 & 0.32\\
ATM & 0.38 & 0.40 & 0.45 & 0.35 & 0.50 & 0.37\\
UBP & 0.52 & 0.28 & 0.37 & 0.40 & 0.47 & 0.33\\
BrainFLORA uni & 0.37 & 0.35 & 0.38 & 0.45 & 0.55 & 0.38\\
Brain-HIVE & 0.42 & 0.27 & 0.19 & 0.32 & 0.39 & 0.28\\
Neural-MCRL & 0.48 & 0.23 & 0.40 & 0.12 & 0.35 & 0.53\\
AlignVis & 0.22 & 0.27 & 0.32 & 0.20 & 0.55 & 0.57\\
HyFI & 0.27 & 0.25 & 0.37 & 0.42 & 0.33 & 0.52\\
NeuroBridge & 0.28 & 0.33 & 0.18 & 0.12 & 0.35 & 0.22\\
ATS & 0.17 & 0.18 & 0.35 & 0.42 & 0.47 & 0.43\\
Visual Blur & 0.07 & 0.23 & 0.27 & 0.25 & 0.50 & 0.52\\
ViEEG & 0.35 & 0.33 & 0.40 & 0.25 & 0.57 & 0.50\\
iTransformer & 0.40 & 0.45 & 0.50 & 0.30 & 0.70 & 0.40\\
\bottomrule\end{tabular*}\par\endgroup

\end{table*}

\paragraph{Performance of the negative controls.}
Under Joint training, AlignVis scores 0.32\% with shuffled EEG and 0.20\% with matched-wrong EEG (Table~\ref{tab:abs-control}). A wrong-stimulus permutation excludes the correct assignment by construction, so its empirical score can fall below uniform chance.
Main Table~\ref{tab:eeg-pairing} reports paired differences, including correct minus matched-wrong. The zero-input result measures behavior on an out-of-distribution signal.

\section{Target Adaptation and Visual Representations}\label{app:adaptation}
\subsection{Independent-Cohort and Test-Cohort Adaptation}
I4 draws calibration images from the training partition. Its comparison with T4 at 800 trials changes both stimulus overlap and the number of sessions represented per image (Table~\ref{tab:adapt}).

\begin{table*}[t]
\SupCaption{table}{\textbf{Unlabeled target adaptation from frozen LOSO checkpoints.} The upper rows define calibration access. The lower rows show Top-1 and paired gains over Frozen with 95\% intervals. Harmed counts participants whose accuracy declined.}{tab:adapt}
{\SupTableStyle{8.1}\begin{tabular*}{\linewidth}{@{\extracolsep{\fill}}llrrrr@{}}\toprule
\textbf{Access} & \textbf{Uses test stimuli?} & \textbf{Images} & \textbf{Repeats/image} & \textbf{Sessions/image} & \textbf{Total trials}\\\midrule
Source only (strict LOSO) & No calibration & 0 & 0 & 0 & 0\\
I4 & No & 200 & 4 & 2 & 800\\
T4 & Yes & 200 & 4 & 4 & 800\\
T80 & Yes & 200 & 80 & 4 & 16,000\\\bottomrule
\end{tabular*}\par}
\smallskip
\begingroup\SupTableStyle{8.1}
\begin{tabular*}{\linewidth}{@{\extracolsep{\fill}}llrrr@{}}\toprule
\textbf{System} & \textbf{Access} & \textbf{Top-1 \ensuremath{\pm} SD (\%)} & \textbf{Gain [95\% interval] (pp)} & \textbf{Harmed / 10}\\\midrule
NICE-BASE & Frozen & 4.35 \ensuremath{\pm} 1.46 & Reference & —\\
 & I4 & 10.56 \ensuremath{\pm} 6.97 & 6.21 [2.60, 10.32] & 2\\
 & T4 & 12.15 \ensuremath{\pm} 7.79 & 7.80 [3.70, 12.39] & 2\\
 & T80 & 14.42 \ensuremath{\pm} 8.72 & 10.07 [5.39, 15.20] & 0\\
\addlinespace[2.5pt]
ATM & Frozen & 5.73 \ensuremath{\pm} 2.65 & Reference & —\\
 & I4 & 10.80 \ensuremath{\pm} 4.23 & 5.06 [3.59, 6.61] & 0\\
 & T4 & 12.76 \ensuremath{\pm} 4.75 & 7.03 [5.28, 8.85] & 0\\
 & T80 & 15.50 \ensuremath{\pm} 5.38 & 9.76 [7.66, 11.90] & 0\\
\addlinespace[2.5pt]
NeuroBridge & Frozen & 7.05 \ensuremath{\pm} 3.47 & Reference & —\\
 & I4 & 11.53 \ensuremath{\pm} 5.96 & 4.48 [2.09, 7.08] & 3\\
 & T4 & 13.16 \ensuremath{\pm} 6.91 & 6.11 [3.21, 9.12] & 2\\
 & T80 & 15.43 \ensuremath{\pm} 8.02 & 8.38 [4.84, 11.92] & 1\\
\addlinespace[2.5pt]
iTransformer & Frozen & 5.99 \ensuremath{\pm} 1.85 & Reference & —\\
 & I4 & 13.74 \ensuremath{\pm} 6.28 & 7.75 [4.75, 11.06] & 1\\
 & T4 & 15.81 \ensuremath{\pm} 6.99 & 9.83 [6.37, 13.52] & 1\\
 & T80 & 18.64 \ensuremath{\pm} 7.80 & 12.66 [8.68, 16.78] & 0\\
 & SATTC & 14.75 \ensuremath{\pm} 6.63 & 8.77 [5.53, 12.27] & 1\\
\bottomrule\end{tabular*}\par\endgroup

\end{table*}

Within each system, fold and seed, all access conditions use the same frozen checkpoint. The test set is fixed across access conditions. Each of the 200 EEG queries averages 80 presentations, and every condition ranks the same 200 gallery images.
Moment matching uses the EEG embedding $z$ before unit normalization. We estimated coordinate-wise moments $(\mu_s,v_s)$ on the source training embeddings and $(\mu_c,v_c)$ on the designated calibration cohort. The transformed embedding is
\[
 \begin{aligned}
 a_j &= \sqrt{(v_{s,j}+\epsilon)/(v_{c,j}+\epsilon)},\\
 \tilde z_j &= a_j(z_j-\mu_{c,j})+\mu_{s,j},\\
 \epsilon &=10^{-4}\operatorname{mean}(v_s)+10^{-8}.
 \end{aligned}
\]
We normalized $\tilde z$ to unit length before scoring it against the fixed gallery. Native SATTC~\citep{evbP08} is an additional test-cohort procedure on the iTransformer backbone with the T80 budget.

\subsection{Subject-Level Gains and Negative Transfer}
Table~\ref{tab:adapt} pairs each calibrated value with the same participant's frozen value. I4 improved mean Top-1 accuracy by 6.21 pp for NICE-BASE, although accuracy declined for two of ten participants. For NeuroBridge the number declining falls from three under I4 to one under T80.

For iTransformer, moment matching reaches 18.64\% Top-1 with T80, compared with 14.75\% for native SATTC on the same test cohort.

\subsection{Visual Targets Under a Common Projection Interface}\label{app:targets}
A native visual representation can change dimensionality as well as the information supplied to the EEG encoder. To keep the EEG output head fixed, we fitted a separate centered, unwhitened 512-component PCA to each visual model's frozen training-image embeddings. We used the training mean and projection for validation and gallery images and normalized each projected embedding. The first two arms use CLIP RN50 and ViT-L/14~\citep{radford2021clip}. The third uses OpenCLIP ViT-H/14~\citep{cherti2023openclip}. Within each EEG system, all arms retain the same EEG architecture, head dimension, training loss and update budget. Each arm fits the EEG mapping to its own projected training targets.

\begin{table*}[t]
\SupCaption{table}{\textbf{Visual targets with a shared PCA512 interface.} Top-1 accuracy (\%) uses CLIP and OpenCLIP targets~\citep{radford2021clip,cherti2023openclip} with PCA fitted on training images. The last column compares the two CLIP arms, L/14 minus RN50, within participant. H/14 uses OpenCLIP. Brackets give the 95\% interval for the mean gain in pp.}{tab:targets}
\begingroup\SupTableStyle{8.8}
\begin{tabular*}{\linewidth}{@{\extracolsep{\fill}}llrrrr@{}}\toprule
\textbf{System} & \textbf{Regime} & \textbf{RN50} & \textbf{L/14} & \textbf{H/14} & \textbf{L/14 minus RN50 (pp)}\\\midrule
NICE-BASE & WS & 12.45 & 14.89 & 16.57 & 2.44 [2.04, 2.96]\\
 & LOSO & 3.98 & 5.00 & 5.73 & 1.02 [0.66, 1.43]\\
\addlinespace[2.5pt]
ATM & WS & 22.48 & 25.78 & 28.27 & 3.31 [2.88, 3.74]\\
 & LOSO & 5.28 & 6.27 & 7.59 & 0.99 [0.74, 1.25]\\
\addlinespace[2.5pt]
NeuroBridge & WS & 26.57 & 30.93 & 32.14 & 4.36 [3.62, 5.17]\\
 & LOSO & 6.41 & 7.78 & 8.46 & 1.37 [0.97, 1.84]\\
\bottomrule\end{tabular*}\par\endgroup

\end{table*}

\paragraph{Visual target ordering.}
All six system--regime combinations in Table~\ref{tab:targets} have the same Top-1 ordering, H/14 above L/14 above RN50. For NeuroBridge, the L/14 gain over RN50 is 4.36 [3.62, 5.17] pp under WS and 1.37 [0.97, 1.84] pp under LOSO. Each teacher bundles an architecture and pretraining history, and each has its own training-fitted PCA projection.

\section{Data Requirements for Retrieval}\label{app:requirements}
\subsection{Test-Trial Averaging and Acquisition Budget}
A test image can contribute one EEG epoch or the average of as many as 80. We varied the number of test repeats with the training set, checkpoint, gallery and source normalization fixed. At 1, 2, 4, 8, 16 and 40 repeats, we averaged performance over eight nested draws per image. The 80-repeat endpoint uses all trials once. For 200 images a single evaluation at budget $r$ uses $200r$ presentations per participant. Repeated draws reuse the available trials.

For ATM under WS, increasing the budget from 16 to 80 repetitions raises Top-1 from 18.85\% to 24.60\%. Across the 200 test images, each participant supplies 12,800 additional presentations for a gain of 5.75 pp (main Figure~\ref{fig:data-requirements}(a)).

\subsection{Counting Images, EEG Pairs, and Updates}
A training pair consists of one participant's EEG response to one photograph, averaged over four presentations. An image viewed by nine participants therefore counts as one photograph and nine pairs. The image-coverage and source-diversity comparisons specify the number of optimizer updates separately because a training pair can be revisited across epochs.

\subsection{Training Image Coverage}
We varied the number of distinct images per concept while retaining all 1,488 training concepts. Each subset is nested in the next and shared across participants. The number of distinct training images is 1,488, 4,464, 7,440 or 14,880. We revisited the smaller sets to match the full-data update budget and validation frequency within each regime.
\begin{supplementwide}
\SupAsset{\includegraphics[width=\linewidth]{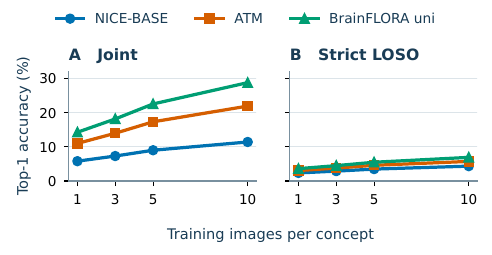}}{\textbf{Retrieval accuracy by number of training images per concept.} Each curve uses one, three, five or ten images per concept and the full-data update budget for its regime.}{fig:resource-curves}
\end{supplementwide}

ATM loses 7.93 pp under Joint training when each concept contributes three images, versus 1.95 pp under LOSO. The contrast measures the value of more visual examples within the same concepts. Removing entire concepts would change the coverage of the training vocabulary as well.

Figure~\ref{fig:resource-curves} compares Joint fitting and strict LOSO at the same four training-image budgets. Three training images per concept give BrainFLORA uni 18.12\% under Joint fitting, compared with 28.72\% using ten images. The paired loss is 10.60 [9.79, 11.48] pp. Under strict LOSO the corresponding means are 4.50\% and 6.91\%. Matching the training steps makes small sets recur more often. The measured change therefore describes exchanging additional views of the same training pairs for observations of previously unseen images within a concept.

\subsection{Gallery Size, Difficulty, and Ranking Stability}
Each reduced gallery retains the positive. Random galleries sample distractors without replacement at $K=2,10,50$, using 20 draws per size. The hard-10 gallery instead takes the nine nearest other test images to the positive in frozen DINOv2 ViT-B/14 space~\citep{oquab2024dinov2}. The evaluator uses the positive ID to construct the gallery without passing that information to the decoder. Hard-10 and full-200 each use one gallery per query. Random Top-1 equals $100/K$ percent.

Brain-HIVE's strict LOSO Top-1 falls from 49.61\% with random-10 to 28.36\% with hard-10. Despite the loss in accuracy, all five systems retain the full-gallery ordering under LOSO with either 10-image gallery (Figure~\ref{fig:gallery}).
\begin{supplementwide}
\SupAsset{\includegraphics[width=\linewidth]{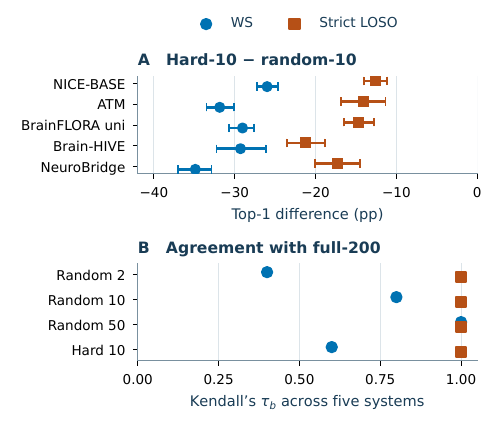}}{\textbf{Gallery difficulty and stability of system rankings.} (A) Participant-paired hard-10 minus random-10 Top-1 change and its 95\% interval, with identical candidate counts. (B) Rank correlation across five systems per gallery and regime against the full-200 ordering. All four strict LOSO comparisons preserve the five-system ranking despite the accuracy loss under hard distractors.}{fig:gallery}
\end{supplementwide}

\subsection{Source-Subject Diversity at Matched Training Budgets}\label{app:sources}
Table~\ref{tab:source-budget} gives the allocations behind Figure~\ref{fig:source-diversity}. For each held-out participant, we generated three orderings of the other nine and took the first 1, 3, 5 or 9 sources from each ordering. Under full allocation, every selected source contributes all 14,880 training images. Under matched allocation, each image is assigned to just one selected source, with four trials averaged for that pair. The assignment stays fixed across epochs. The matched series uses the same number of paired observations and the nine-source update schedule at every source count. Only eligible sources supply normalization and validation observations.
\begin{supplementwide}
\SupAsset{\includegraphics[width=\linewidth]{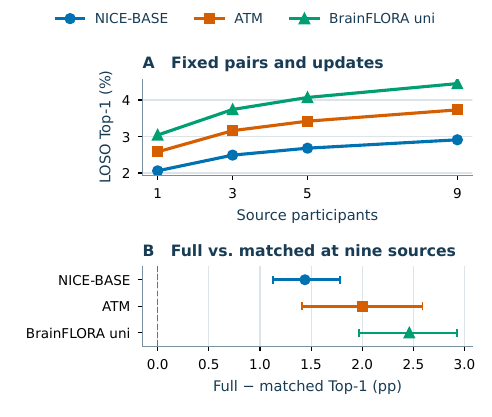}}{\textbf{Source-participant diversity under strict LOSO.} (A) Mean Top-1 at one, three, five and nine sources with matched pairs and updates. (B) Participant-paired full-minus-matched Top-1 differences at nine sources, with 95\% intervals. Both nine-source arms use the same participants and update schedule.}{fig:source-diversity}
\end{supplementwide}

\begin{table*}[t]
\SupCaption{table}{\textbf{Training budgets by source-participant count.} Counts are per fit. All arms use 14,880 distinct images. At one source the observations coincide and scheduled updates differ. At nine sources the updates coincide and the number of distinct pairs differs.}{tab:source-budget}
\begingroup\SupTableStyle{8.8}
\begin{tabular*}{\linewidth}{@{\extracolsep{\fill}}rrrrrrr@{}}\toprule
\multicolumn{1}{c}{\textbf{}} & \multicolumn{3}{c}{\textbf{Full}} & \multicolumn{3}{c}{\textbf{Matched}}\\
\cmidrule(lr){2-4}\cmidrule(lr){5-7}
\textbf{Sources} & \textbf{Pairs} & \textbf{Trials} & \textbf{Updates} & \textbf{Pairs} & \textbf{Trials} & \textbf{Updates}\\\midrule
1 & 14,880 & 59,520 & 23,400 & 14,880 & 59,520 & 210,600\\
3 & 44,640 & 178,560 & 70,200 & 14,880 & 59,520 & 210,600\\
5 & 74,400 & 297,600 & 117,000 & 14,880 & 59,520 & 210,600\\
9 & 133,920 & 535,680 & 210,600 & 14,880 & 59,520 & 210,600\\
\bottomrule\end{tabular*}\par\endgroup

\end{table*}

From one to nine sources, NICE-BASE gains 1.52 pp with full allocation and 0.85 [0.67, 1.04] pp with matched allocation. The matched gains are 1.15 [0.81, 1.50] pp for ATM and 1.41 [1.13, 1.68] pp for BrainFLORA uni. Additional participants can help even when no new paired observations are added. 
At nine sources the two allocation rules use the same subjects and the same number of updates. They differ in whether the decoder sees multiple participants responding to each training image. ATM's full-minus-matched difference there is 2.00 [1.41, 2.59] pp, with a Top-5 difference of 5.97 [4.51, 7.34] pp. These endpoints are the cleanest comparison of repeated cross-participant coverage of an image under this schedule.

\section{Temporal, Spatial, and Reliability Analyses}\label{app:reliability}
\subsection{Temporal Occlusion and Short-Window Retraining}
A frozen encoder trained on one-second inputs can fail on a short prefix both because evidence was removed and because the masked input is unfamiliar. We evaluated the frozen model on the first 100, 200, 300, 500, 750 or 1,000 ms of each query. A second test masked one contiguous 200-ms block of the full epoch. In both tests we zeroed the excluded samples after source normalization. For the 200- and 500-ms prefixes, we also trained and evaluated new models using the same prefix.
\begin{supplementwide}
\SupAsset{\includegraphics[width=\linewidth]{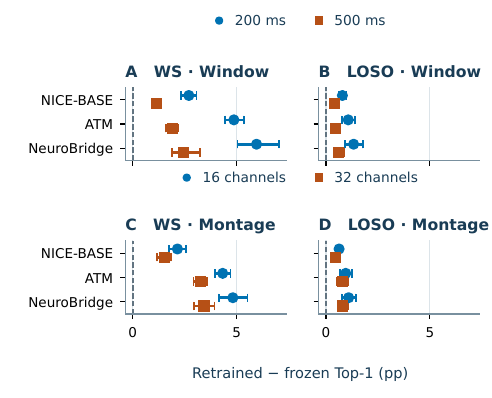}}{\textbf{Retraining gains for shorter EEG windows and reduced channel sets.} Each pair compares retrained and frozen models at the same test input budget. Points show the ten-subject mean Top-1 gain and bars show the paired 95\% interval.}{fig:restricted}
\end{supplementwide}

\begin{supplementwide}
\SupAsset{\includegraphics[width=\linewidth]{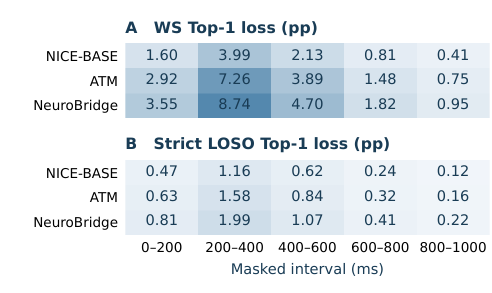}}{\textbf{Retrieval loss after masking a 200-ms EEG interval.} Each cell gives the participant-paired Top-1 difference (full minus masked, pp) for a frozen decoder. Darker blue indicates a larger loss on the shared WS and strict LOSO scale.}{fig:blocks}
\end{supplementwide}

For NICE-BASE under WS, retraining on 200 ms gains 2.70 [2.32, 3.07] pp over the frozen 200-ms condition, but remains 4.07 [3.29, 4.90] pp below full input.
Figure~\ref{fig:blocks} shows ATM's largest WS loss after removal of 200--400 ms (7.26 [6.52, 8.05] pp). All masks are applied offline to the resampled, onset-aligned epoch.

\subsection{Regional Occlusion and Sparse-Montage Retraining}
We compared each regional mask with 20 random masks of equal size (Figure~\ref{fig:regions}). The parietal mask contained 15 electrodes, and each of the other four masks contained 12. Occipital removal gives the largest excess WS loss for all three systems (1.94--4.17 pp). The masks localize sensitivity to scalp channels. Volume conduction limits anatomical interpretation.
\begin{supplementwide}
\SupAsset{\includegraphics[width=\linewidth]{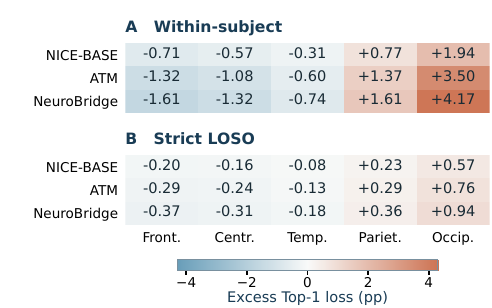}}{\textbf{Retrieval loss after masking an electrode region.} Each cell gives excess Top-1 loss (pp) relative to 20 equal-size random masks. Positive values indicate greater loss for the named region and negative values greater loss for the random masks. Both panels use the same zero-centered scale.}{fig:regions}
\end{supplementwide}

On the same 16-channel WS input, NICE-BASE gains 2.15 [1.73, 2.58] pp through retraining (Figure~\ref{fig:restricted}). Its retrained accuracy is 11.04\%, compared with 13.75\% at 63 channels.

\subsection{Robustness to Input Perturbations}
We perturbed the 80-repeat average after applying the frozen source normalization. Gaussian noise uses an RMS reference from source-training mean4 data. The 20, 10 and 0 dB conditions scale a shared random noise field. Channel dropout removes 6, 16 or 32 channels using nested permutations, shared across queries within a participant and draw. Noise and dropout each use five draws per level. Gain multiplies the normalized query by 0.5 or 2.

\begin{table*}[t]
\SupCaption{table}{\textbf{Top-1 accuracy (\%) under EEG perturbations.} Noise columns use source-reference SNR. Drop columns give the number of channels removed. The final columns scale the signal amplitude.}{tab:stress}
\begingroup\SupTableStyle{8.5}
\begin{tabular*}{\linewidth}{@{\extracolsep{\fill}}llrrrrrrrrr@{}}\toprule
\multicolumn{3}{c}{\textbf{}} & \multicolumn{3}{c}{\textbf{Noise}} & \multicolumn{3}{c}{\textbf{Channel dropout}} & \multicolumn{2}{c}{\textbf{Gain}}\\
\cmidrule(lr){4-6}\cmidrule(lr){7-9}\cmidrule(lr){10-11}
\textbf{System} & \textbf{Regime} & \textbf{Clean} & \textbf{20 dB} & \textbf{10 dB} & \textbf{0 dB} & \textbf{6} & \textbf{16} & \textbf{32} & \textbf{×0.5} & \textbf{×2}\\\midrule
NICE-BASE & WS & 13.75 & 13.34 & 11.36 & 6.33 & 12.82 & 10.83 & 7.26 & 12.70 & 12.31\\
 & LOSO & 4.35 & 4.23 & 3.66 & 2.19 & 4.08 & 3.50 & 2.46 & 4.04 & 3.93\\
\addlinespace[2.5pt]
ATM & WS & 24.60 & 23.88 & 20.27 & 11.11 & 22.92 & 19.31 & 12.80 & 22.52 & 21.80\\
 & LOSO & 5.73 & 5.58 & 4.80 & 2.81 & 5.37 & 4.59 & 3.17 & 5.27 & 5.11\\
\addlinespace[2.5pt]
BrainFLORA uni & WS & 19.53 & 18.94 & 16.10 & 8.87 & 18.19 & 15.34 & 10.20 & 17.99 & 17.42\\
 & LOSO & 6.91 & 6.71 & 5.75 & 3.32 & 6.46 & 5.50 & 3.77 & 6.39 & 6.19\\
\addlinespace[2.5pt]
Brain-HIVE & WS & 20.04 & 19.48 & 16.57 & 9.12 & 18.72 & 15.78 & 10.49 & 18.46 & 17.88\\
 & LOSO & 10.21 & 9.93 & 8.48 & 4.78 & 9.55 & 8.09 & 5.46 & 9.44 & 9.15\\
\addlinespace[2.5pt]
NeuroBridge & WS & 29.37 & 28.49 & 24.18 & 13.21 & 27.35 & 23.02 & 15.23 & 27.25 & 26.39\\
 & LOSO & 7.05 & 6.84 & 5.87 & 3.38 & 6.59 & 5.61 & 3.84 & 6.54 & 6.35\\
\bottomrule\end{tabular*}\par\endgroup

\end{table*}

In Table~\ref{tab:stress}, 0 dB noise reduces ATM's WS accuracy from 24.60\% to 11.11\%. Dropping 32 channels leaves 12.80\%.

\subsection{Confidence Calibration and Selective Retrieval}\label{app:confidence}
We use NICE-BASE as a case study because its candidate scores, correct answers and fitted temperatures allow both probability mappings to be checked for every query. For native logits $z$ and candidate $j$, $p_j=\exp(z_j/T)/\sum_k\exp(z_k/T)$. The native arm uses $T=1$. The scaled arm fits $10^{-3}\leq T\leq10^3$ per checkpoint by minimizing source-balanced negative log likelihood on the 1,660-image validation gallery~\citep{guo2017calibration}. Confidence is $\max_j p_j$.
\begin{supplementwide}
\SupAsset{\includegraphics[width=\linewidth]{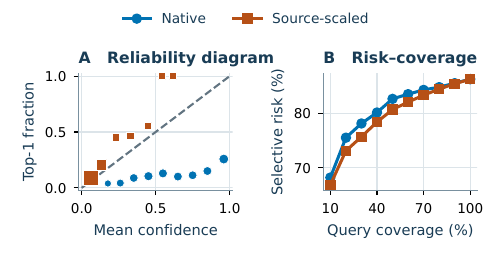}}{\textbf{Confidence calibration and selective risk for NICE-BASE under WS.} (A) Ten equal-width confidence bins pooled over ten participants and three seeds. Marker area reflects occupancy. Reported ECE averages the individual participant--seed ECE values. (B) Ten operating points from the same evaluations. Risk is averaged over three seeds per participant and then over ten participants. AURC uses all 200 coverage levels.}{fig:calibration}
\end{supplementwide}

\paragraph{Calibration and retrieval order.}
For WS, scaling reduces mean ECE from 0.5070 to 0.0534 and AURC from 0.7883 to 0.7745 (Figure~\ref{fig:calibration}). Positive temperature preserves candidate order within a query. The maximum probability can reorder confidence across queries, changing which errors are retained.

\paragraph{Selective risk.}
Within each participant and seed, we ranked the 200 queries by descending confidence and averaged error within exact confidence ties at the retention boundary. At each coverage, risk is the expected error rate among the retained queries~\citep{geifman2017selective}. A source-fixed threshold instead applies the validation cutoff for 80\% retention, including ties, to each test query. For scaled WS predictions, the source cutoff retains 100\% of test queries, while risk among the most confident 80\% is 84.44\%.

\begin{figure*}[t]
\SupAsset{\includegraphics[width=\linewidth]{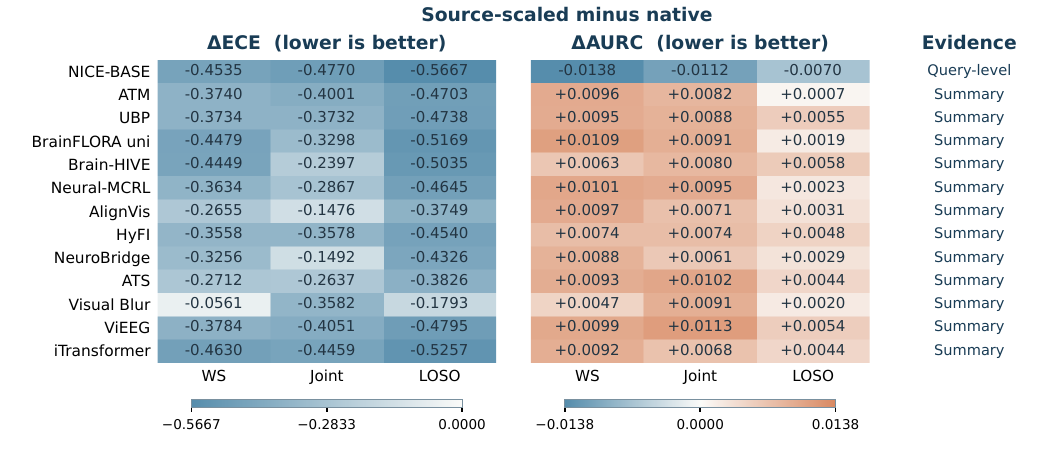}}{\textbf{Temperature scaling across 13 retrieval systems.} Panels report source-scaled minus native expected calibration error (ECE) and area under the risk--coverage curve (AURC). Colors encode the changes in each metric. The evidence column indicates whether the values were checked against individual evaluations (NICE-BASE) or obtained from system summaries (the other 12).}{fig:calibration-changes}
\end{figure*}

Across the 13 systems, ECE falls in all 39 system--regime pairs and AURC rises in 36 (Figure~\ref{fig:calibration-changes}). For ATM under WS, the ECE improvement shown in main Figure~\ref{fig:sensitivity-reliability}(b) accompanies higher selective risk. Its reported ECE falls from 0.5248 to 0.1508 while AURC rises from 0.6894 to 0.6990. Table~\ref{tab:calibration-metrics} gives the other three metrics.

\begin{table*}[t]
\SupCaption{table}{\textbf{Calibration and selective risk after temperature scaling.} Values give source-scaled minus native NLL, Brier score and risk at 80\% coverage. Negative values indicate a reduction. The source for each system is identified in Figure~\ref{fig:calibration-changes}. The supplementary data include participant-paired 95\% intervals for all five metrics.}{tab:calibration-metrics}
\begingroup\SupTableStyle{8.3}
\begin{tabular*}{\linewidth}{@{\extracolsep{\fill}}lrrrrrrrrr@{}}\toprule
\multicolumn{1}{c}{\textbf{}} & \multicolumn{3}{c}{\textbf{NLL}} & \multicolumn{3}{c}{\textbf{Brier}} & \multicolumn{3}{c}{\textbf{Risk at 80\% coverage (pp)}}\\
\cmidrule(lr){2-4}\cmidrule(lr){5-7}\cmidrule(lr){8-10}
\textbf{System} & \textbf{WS} & \textbf{Joint} & \textbf{LOSO} & \textbf{WS} & \textbf{Joint} & \textbf{LOSO} & \textbf{WS} & \textbf{Joint} & \textbf{LOSO}\\\midrule
NICE-BASE & -2.9174 & -3.0991 & -5.9416 & -0.3190 & -0.3369 & -0.4413 & -0.35 & -0.17 & -0.23\\
ATM & -1.1057 & -1.2625 & -2.2393 & -0.2715 & -0.2992 & -0.4395 & +0.04 & +0.17 & +0.04\\
UBP & -1.1581 & -1.1364 & -2.0370 & -0.2704 & -0.2647 & -0.4116 & +0.15 & -0.06 & +0.02\\
BrainFLORA uni & -1.3877 & -0.9768 & -2.3393 & -0.3358 & -0.2347 & -0.4537 & +0.12 & +0.10 & +0.04\\
Brain-HIVE & -1.3865 & -0.6662 & -2.0069 & -0.3307 & -0.1639 & -0.4203 & +0.06 & +0.13 & +0.06\\
Neural-MCRL & -1.0804 & -0.8023 & -2.3006 & -0.2542 & -0.1985 & -0.4436 & +0.00 & +0.08 & -0.02\\
AlignVis & -0.7345 & -0.4118 & -1.7960 & -0.1781 & -0.0878 & -0.3541 & +0.10 & +0.13 & +0.02\\
HyFI & -1.0591 & -1.1129 & -1.9745 & -0.2536 & -0.2595 & -0.3965 & +0.10 & +0.06 & +0.15\\
NeuroBridge & -0.9334 & -0.4223 & -2.0140 & -0.2288 & -0.0926 & -0.4026 & +0.12 & +0.19 & +0.02\\
ATS & -0.7428 & -0.8086 & -1.7762 & -0.1803 & -0.1757 & -0.3447 & +0.19 & +0.15 & +0.04\\
Visual Blur & -0.1506 & -1.0380 & -1.0060 & -0.0274 & -0.2571 & -0.1792 & +0.10 & +0.06 & -0.04\\
ViEEG & -1.1806 & -1.2657 & -2.1443 & -0.2733 & -0.2980 & -0.4318 & +0.08 & +0.21 & +0.02\\
iTransformer & -1.4285 & -1.4730 & -2.3510 & -0.3504 & -0.3345 & -0.4646 & +0.08 & +0.15 & -0.00\\
\bottomrule\end{tabular*}\par\endgroup

\end{table*}

\section{Additional Tasks and Scope}\label{app:additional}
\subsection{EEG-Conditioned Image Reconstruction and Generator Controls}\label{app:generation}
Three Joint systems produce images from EEG queries.

\begin{table*}[t]
\SupCaption{table}{\textbf{Image reconstruction and generator controls under Joint training.} The upper rows give correct EEG, matched-wrong EEG and prior-only performance. Higher PixCorr, SSIM~\citep{wang2004ssim} and two-way identification (ID) and lower LPIPS~\citep{zhang2018lpips} favor the reference. Chance ID is 50\%. The lower rows give differences between correct EEG and each control with participant-paired 95\% intervals.}{tab:generation}
\begingroup\SupTableStyle{8.8}
\begin{tabular*}{\linewidth}{@{\extracolsep{\fill}}llrrrrr@{}}\toprule
\textbf{System} & \textbf{Input} & \textbf{PixCorr} & \textbf{SSIM} & \textbf{LPIPS} & \textbf{CLIP ID (\%)} & \textbf{DINO ID (\%)}\\\midrule
ATM & correct & 0.1755 & 0.2272 & 0.6667 & 72.15 & 68.58\\
 & matched wrong & 0.0214 & 0.1028 & 0.8175 & 50.38 & 50.59\\
 & prior only & 0.0222 & 0.1042 & 0.8087 & 50.99 & 50.51\\
\addlinespace[2.5pt]
Brain-HIVE & correct & 0.2047 & 0.2470 & 0.6243 & 76.66 & 73.32\\
 & matched wrong & 0.0213 & 0.1039 & 0.8155 & 50.30 & 50.73\\
 & prior only & 0.0236 & 0.1071 & 0.8090 & 50.67 & 50.67\\
\addlinespace[2.5pt]
BrainFLORA uni & correct & 0.1986 & 0.2395 & 0.6379 & 74.98 & 71.86\\
 & matched wrong & 0.0192 & 0.1034 & 0.8191 & 50.53 & 50.40\\
 & prior only & 0.0206 & 0.1058 & 0.8145 & 51.03 & 50.15\\
\bottomrule\end{tabular*}\par\endgroup

\medskip
\begingroup\SupTableStyle{8.4}
\begin{tabular*}{\linewidth}{@{\extracolsep{\fill}}llrrr@{}}\toprule
\textbf{System} & \textbf{Control} & \textbf{CLIP gain (pp)} & \textbf{DINO gain (pp)} & \textbf{LPIPS difference}\\\midrule
ATM & matched wrong & +21.77 [19.98, 23.43] & +17.99 [16.53, 19.31] & -0.1508 [-0.1587, -0.1429]\\
 & prior only & +21.16 [18.94, 23.07] & +18.07 [16.39, 19.71] & -0.1419 [-0.1498, -0.1342]\\
\addlinespace[2.5pt]
Brain-HIVE & matched wrong & +26.36 [23.96, 28.76] & +22.59 [20.65, 24.70] & -0.1912 [-0.2030, -0.1807]\\
 & prior only & +25.99 [23.62, 28.47] & +22.66 [20.66, 24.83] & -0.1847 [-0.1950, -0.1751]\\
\addlinespace[2.5pt]
BrainFLORA uni & matched wrong & +24.45 [22.54, 26.48] & +21.46 [19.72, 23.25] & -0.1812 [-0.1913, -0.1717]\\
 & prior only & +23.94 [21.56, 26.54] & +21.71 [19.81, 23.60] & -0.1766 [-0.1909, -0.1635]\\
\bottomrule\end{tabular*}\par\endgroup

\end{table*}

\begin{table*}[t]
\SupCaption{table}{\textbf{Visual-target configurations in Brain-HIVE.} Under Joint training, the first six rows for Brain-HIVE~\citep{evbP07} vary the visual targets. Capacity is the recorded percentage relative to the full model. The last row gives the main retrieval and correct-EEG reconstruction reference.}{tab:fusion}
\begingroup\SupTableStyle{8.8}
\begin{tabular*}{\linewidth}{@{\extracolsep{\fill}}lrrrrr@{}}\toprule
\textbf{Targets} & \textbf{Capacity (\%)} & \textbf{Top-1 \ensuremath{\pm} SD (\%)} & \textbf{LPIPS} & \textbf{CLIP ID (\%)} & \textbf{DINO ID (\%)}\\\midrule
B16 & 65.0 & 30.34 \ensuremath{\pm} 5.06 & 0.6374 & 73.76 & 71.46\\
B32 & 65.0 & 27.60 \ensuremath{\pm} 4.37 & 0.6472 & 72.69 & 69.95\\
VAE & 65.0 & 23.07 \ensuremath{\pm} 3.71 & 0.6730 & 69.98 & 67.93\\
B32 + VAE & 85.0 & 30.59 \ensuremath{\pm} 4.91 & 0.6415 & 74.28 & 71.55\\
B16 + VAE & 85.0 & 33.37 \ensuremath{\pm} 5.59 & 0.6312 & 75.82 & 72.90\\
B16 + B32 & 85.0 & 34.62 \ensuremath{\pm} 5.55 & 0.6252 & 76.32 & 73.61\\
\midrule
Full (main reference) & 100.0 & 35.60 \ensuremath{\pm} 5.89 & 0.6243 & 76.66 & 73.32\\
\bottomrule\end{tabular*}\par\endgroup

\end{table*}

We generated two images per EEG query, evaluated each image separately, and averaged their scores before aggregating across seeds and participants. Corresponding input conditions used the same sampling seeds within each system. Correct, shuffled, matched-wrong and zero EEG enter through the same conditioning path. Prior-only removes all EEG-derived conditioning, including intermediate features. Substituting zeros at the input still allows the conditioning network to return a nonzero embedding.

The pixel metrics compare resized $256\times256$ RGB images. PixCorr is the correlation of flattened images, SSIM averages local structural similarity across color channels~\citep{wang2004ssim}, and LPIPS uses AlexNet features~\citep{zhang2018lpips}. For two-way identification, compare the reconstruction's cosine similarity to the correct stimulus and to each of the other 199 stimuli in turn, awarding half credit to a tie. The CLIP and DINOv2 encoders~\citep{radford2021clip,oquab2024dinov2} provide two feature spaces for this comparison.

In Table~\ref{tab:generation}, Brain-HIVE gains 26.36 [23.96, 28.76] pp in CLIP identification over matched-wrong EEG and 25.99 [23.62, 28.47] pp over prior-only. Both comparisons hold the generator and its sampling seeds fixed. The supplementary data also include zero and shuffled EEG.

\subsection{Visual-Target Configurations in Brain-HIVE}\label{app:fusion}
In Table~\ref{tab:fusion}, B16 is SynCLR ViT-B/16~\citep{tian2024synclr} and VAE denotes the latent target. The configuration runs identify B32 as SynCLR B/32, whereas the main model uses CLIP ViT-B/32. Table~\ref{tab:fusion} lists the main model separately as a reference.

\subsection{Population and Stimulus Coverage}
Subject intervals describe ten participants on a fixed set of 200 photographs from THINGS-EEG2. Strict LOSO measures transfer to another participant under the same acquisition protocol. Within-concept image variation and transfer to a different recording protocol require additional data. Each training and validation image was presented twice in each of two sessions. A fold with one presentation from each of three training sessions cannot be formed from these observations. Cross-session retrieval remains an open question.

\end{multicols}
\begin{supplementwide}
\pdfbookmark[1]{Qualitative retrieval gallery}{supp-gallery}
\SupAsset{\includegraphics[width=\linewidth]{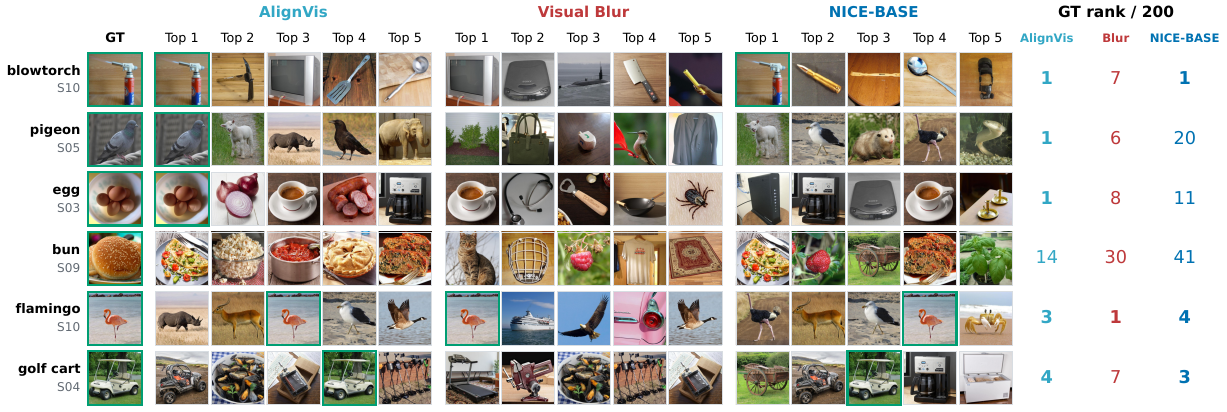}}{\textbf{Joint retrieval for items 7--12 of the prespecified draw.} Each row shows the target, three Top-5 lists and the target ranks among 200 candidates. The rows continue main Figure~\ref{fig:retrieval-examples} in their original order. For S09's bun, AlignVis, Visual Blur and NICE-BASE rank the target 14th, 30th and 41st among 200 images.}{fig:supp-continuation}
\end{supplementwide}

\begin{supplementwide}
\SupAsset{\includegraphics[width=\linewidth]{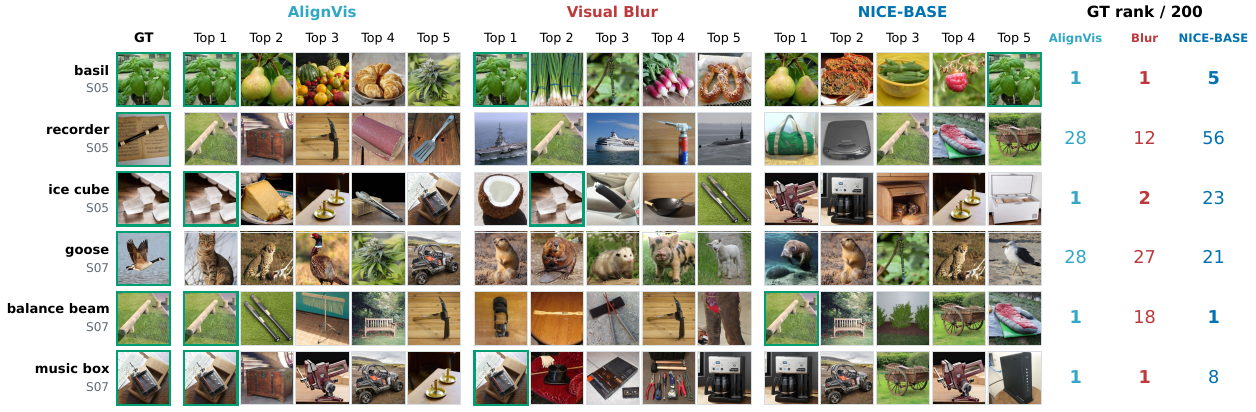}}{\textbf{Joint retrieval for S05 and S07.} Each row shows one Joint seed-0 query. The left column gives the ground-truth image (GT). Each system ranks five candidates from left to right, with the target outlined in green. The columns on the right give the target's rank among all 200 candidates.}{fig:supp-examples}
\end{supplementwide}

\begin{supplementwide}
\SupAsset{\includegraphics[width=\linewidth]{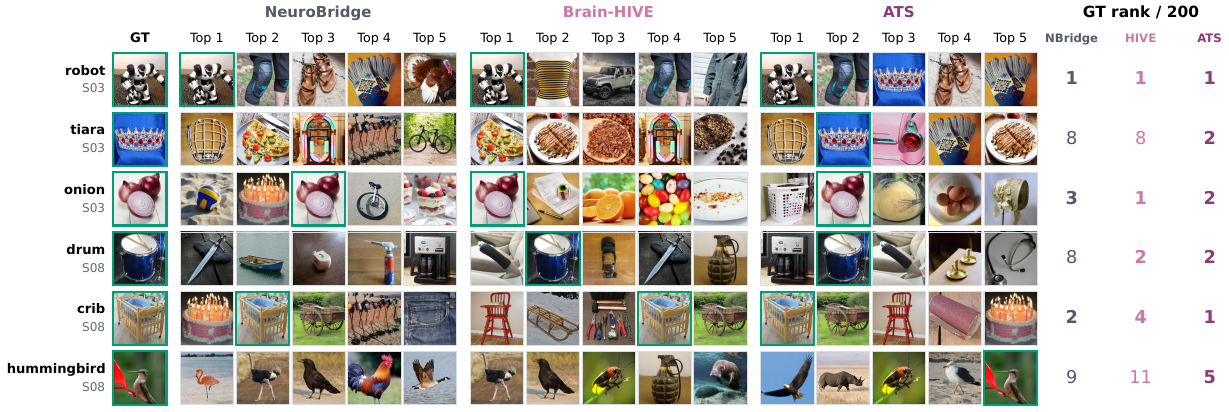}}{\textbf{Joint retrieval for S03 and S08.}}{fig:supp-ranks-02-04}
\end{supplementwide}

\begin{supplementwide}
\SupAsset{\includegraphics[width=\linewidth]{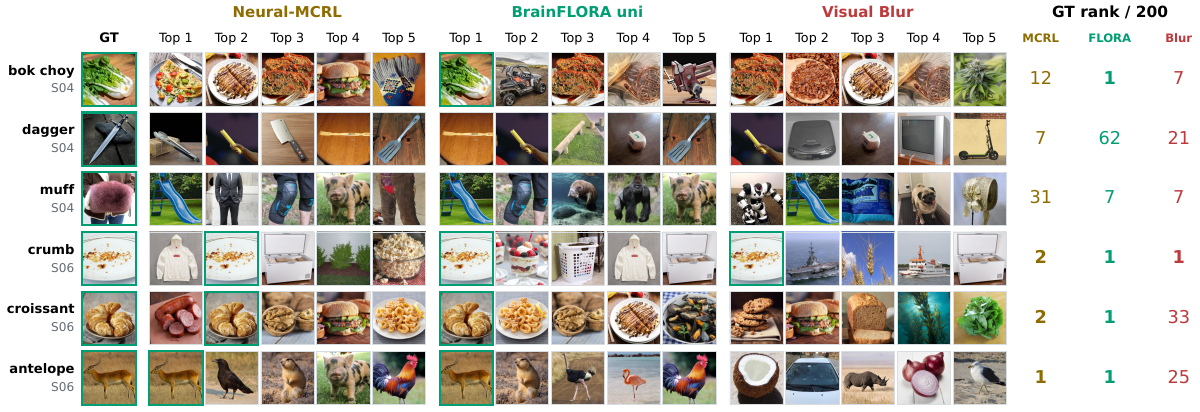}}{\textbf{Joint retrieval for S04 and S06.}}{fig:supp-ranks-05-07}
\end{supplementwide}

\begin{supplementwide}
\SupAsset{\includegraphics[width=\linewidth]{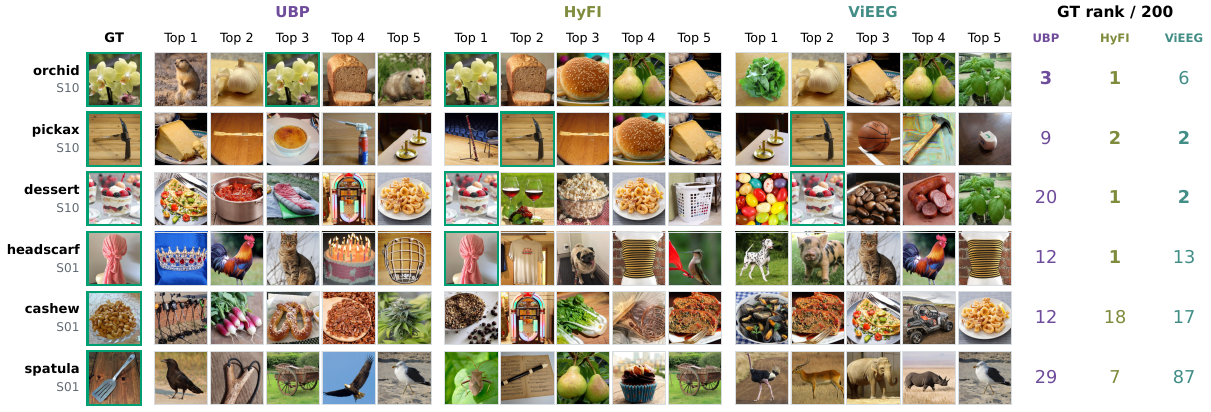}}{\textbf{Joint retrieval for S10 and S01.}}{fig:supp-ranks-08-10}
\end{supplementwide}

\begin{supplementwide}
\SupAsset{\includegraphics[width=\linewidth]{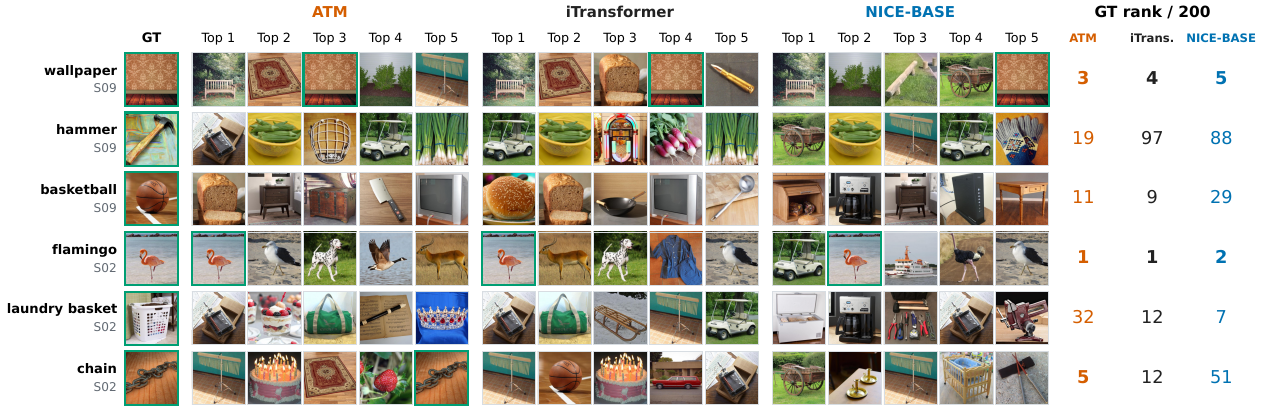}}{\textbf{Joint retrieval for S09 and S02.}}{fig:supp-ranks-11-13}
\end{supplementwide}

\end{document}